\documentclass[10pt,twocolumn,letterpaper]{article}

\usepackage[pagenumbers]{cvpr} % To force page numbers, e.g. for an arXiv version

\usepackage{multirow}
\usepackage{ulem} 
\usepackage{colortbl}
\usepackage{empheq}
\usepackage{marvosym}
\definecolor{cvprblue}{rgb}{0.21,0.49,0.74}
\usepackage[pagebackref,breaklinks,colorlinks,allcolors=cvprblue]{hyperref}

\def\paperID{11165} % *** Enter the Paper ID here
\def\confName{CVPR}
\def\confYear{2026}

\title{Towards Metric-Aware Multi-Person Mesh Recovery by Jointly Optimizing Human Crowd in Camera Space}

\author{
Kaiwen Wang$^{1*}$ \quad
Kaili Zheng$^{1*}$ \quad
Yiming Shi$^{1}$ \quad
Chenyi Guo$^{1}$\textsuperscript{\Letter} \quad
Ji Wu$^{1,2,3}$\textsuperscript{\Letter} \\
$^1$Department of Electronic Engineering, Tsinghua University\\
$^2$College of AI, Tsinghua University\\
$^3$Beijing National Research Center for Information Science and Technology\\
{$^*$ Equal contribution \textsuperscript{\Letter} Corresponding author}\\
{\tt\small \{wkw23, zkl25, shiym23\}@mails.tsinghua.edu.cn} \\
{\tt\small \{guochy, wuji\_ee\}@mail.tsinghua.edu.cn}
}

\begin{document}

\twocolumn[
\begin{center}
    \maketitle
    \vspace{-1.5em}
    {\includegraphics[width=0.9\textwidth]{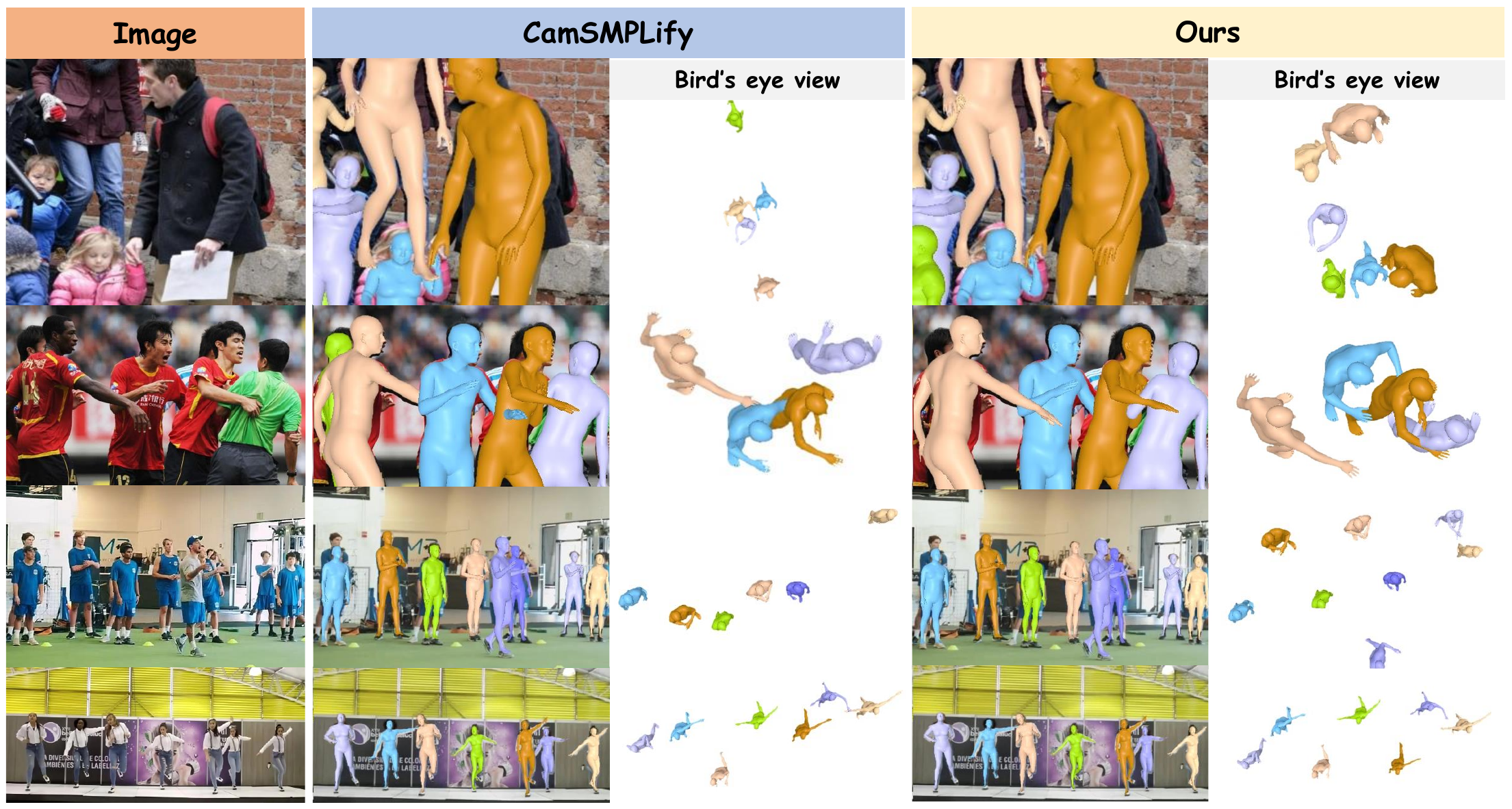}}
    % \captionof{figure}{CamSMPLify \cite{camerahmr} generates pseudo ground-truths by optimizing each individual independently based on 2D cues. However, this per-person optimization often results in unrealistic inter-person spatial relations and inaccurate height estimations. Our Depth-conditioned Translation Optimization (DTO) method addresses this by formulating a joint Maximum a posteriori (MAP) problem. It finds the globally optimal camera-space translations for all individuals by maximizing a posterior probability based on human height prior and depth cues, ensuring a coherent scene reconstruction.}
    \captionof{figure}{CamSMPLify \cite{camerahmr} fits each person independently, causing height and spatial inconsistencies. Our DTO jointly optimizes human translations from height priors and depth cues, ensuring a coherent scene reconstruction.}
    \label{fig:teaser}
\end{center}
]

\begin{abstract}
Multi-person human mesh recovery from a single image is a challenging task, hindered by the scarcity of in-the-wild training data. Prevailing in-the-wild human mesh pseudo-ground-truth (pGT) generation pipelines are single-person-centric, where each human is processed individually without joint optimization. This oversight leads to a lack of scene-level consistency, producing individuals with conflicting depths and scales within the same image. To address this, we introduce \textbf{D}epth-conditioned \textbf{T}ranslation \textbf{O}ptimization (DTO), a novel optimization-based method that jointly refines the camera-space translations of all individuals in a crowd. By leveraging anthropometric priors on human height and depth cues from a monocular depth estimator, DTO solves for a scene-consistent placement of all subjects within a principled Maximum a posteriori (MAP) framework. Applying DTO to the 4D-Humans dataset, we construct \textbf{DTO-Humans}, a new large-scale pGT dataset of 0.56M high-quality, scene-consistent multi-person images, featuring dense crowds with an average of 4.8 persons per image. Furthermore, we propose \textbf{Metric-Aware HMR}, an end-to-end network that directly estimates human mesh and camera parameters in metric scale. This is enabled by a camera branch and a relative metric loss that enforces plausible relative scales. Extensive experiments demonstrate that our method achieves state-of-the-art performance on relative depth reasoning and human mesh recovery.
Code is available at: https://github.com/gouba2333/MA-HMR.
\end{abstract}    
\section{Introduction}
\label{sec:intro}

Recovering 3D human pose and shape from a single image is a fundamental challenge in computer vision with profound implications for applications such as autonomous driving \cite{autonomousdriving}, augmented and virtual reality \cite{ar}, sports analysis \cite{sports}. While much progress has focused on individuals, real-world scenes are typically composed of multiple people. Challenges like heavy occlusions and depth ambiguities prevalent in multi-person scenes make holistic scene understanding a hard problem to solve.

While the construction of in-the-wild 3D human datasets is crucial for training deep learning models, acquiring such data at a large scale remains exceptionally difficult. To overcome this, the community has widely adopted a paradigm of pseudo-ground-truth (pGT) generation \cite{spin, eft, neuralannot, camerahmr}. However, current pipelines are single-person-centric. They optimize each individual in isolation and simply assemble the results. This lack of joint optimization ignores scene-level spatial relationships, yielding physically implausible layouts with incorrect depth ordering and inconsistent relative scales (see Fig.~\ref{fig:teaser}, mid). These inconsistent pGTs limit the potential of powerful end-to-end regression frameworks \cite{psvt, aios, multihmr, sathmr} which are capable of learning complex human-human and human-environment relations, and thus place a data bottleneck in the field of multi-person mesh recovery.

To resolve scene-wide inconsistencies and put people in their place, we model the problem by mirroring human-like inference, which relies on two key information sources: prior knowledge of human height and visual cues of relative depth. We leverage off-the-shelf models \cite{mivolov2} to predict age and gender, from which we derive a statistically-informed height distribution for each individual \cite{height, growth}. For the visual cues, we utilize a monocular depth estimator \cite{depthv2} to extract rich geometric information. We then propose a unified framework to integrate these information, named \textbf{D}epth-conditioned \textbf{T}ranslation \textbf{O}ptimization (DTO). DTO solves a joint Maximum a posteriori (MAP) problem, where the height distribution acts as the prior. The likelihood term is conditioned on two robust depth constraints: inter-human depth relations for positioning and intra-human depth scale for structural consistency. The resulting optimization finds the globally optimal human translations that make the entire crowd spatially consistent in the scene (see Fig.~\ref{fig:teaser}, right).

We then address the data bottleneck by applying DTO to 2M images from the 4D-Humans dataset \cite{instavariety, aic, coco, mpii} and leverage the posterior probability derived from our DTO framework to filter out poorly optimized or inconsistent scenes. This yields \textbf{DTO-Humans}, a large-scale pGT dataset comprising 0.56M high-quality, scene-consistent multi-person images. With 2.7M person instances in dense scenes (avg. 4.8 persons/image), this dataset provides consistent, physically-grounded supervision that has long been a missing piece in training robust multi-person models.

Building upon this dataset, we further propose Metric-Aware HMR, an end-to-end network that directly regresses human mesh parameters in metric scale. A critical challenge in metric recovery is the unknown camera intrinsic parameters, which are intricately coupled with the perceived scale of individuals and the overall scene composition. An end-to-end architecture is uniquely suited to resolve this ambiguity, as it can process global image information. Our network materializes this insight by incorporating a camera branch with the HMR model and simultaneously predicts the camera's field of view (FoV) and human mesh parameters. To guide this joint prediction, we introduce a novel relative metric loss, which explicitly penalizes implausible real-world size relationships among individuals. By learning to predict all parameters concurrently from the full image context, our model learns the complex interplay between scene geometry and individual human scales.

In summary, our contributions are as follows:
\begin{itemize}
\item We propose DTO, a novel MAP optimization framework that jointly optimizes multi-person camera-space translations using both inter-human and intra-human depth constraints to ensure scene-level consistency.

\item We construct DTO-Humans, a large-scale pGT dataset with high scene consistency, paving the way for training more robust multi-person models.

\item We design an end-to-end network, Metric-Aware HMR, featuring a dedicated camera branch and a relative metric loss to achieve true metric-scale mesh recovery.

\item Extensive experiments show that our method achieves state-of-the-art performance on multi-person 3D reconstruction benchmarks, including RH \cite{bev}, 3DPW \cite{3dpw}, CMU-Panoptic \cite{panoptic}, Hi4D \cite{hi4d}, and MuPoTS \cite{mupots}.
\end{itemize}

\section{Related work}
\label{sec:relatedwork}

\subsection{Multi-person Human Mesh Recovery}
Multi-person human mesh recovery aims to regress all human meshes from a single RGB image. Existing methods can be categorized into multi-stage and one-stage approaches. Multi-stage approaches \cite{cliff, 4dhumans, smpler, camerahmr} detect all humans in the image and apply a single-person HMR model to each of them separately. In contrast, single-stage paradigms process the entire image at once to jointly recover all individuals, which allow the model to capture human-human and human-environment interactions. Early methods like ROMP \cite{romp} and BMP \cite{bmp} rely on person-center heatmaps. More recently, DETR \cite{detr, dn, dab}-based frameworks \cite{psvt, aios, multihmr, sathmr} have shown strong performance by using human queries to decode mesh parameters from feature maps. While these end-to-end models have the potential for holistic scene understanding, they highly depend on the quantity and quality of training data.

\subsection{Multi-person pGT Generation}
The scarcity of in-the-wild 3D human data has made pseudo-ground-truth (pGT) generation a cornerstone of modern HMR \cite{eft, spin, neuralannot, prohmr, camerahmr}. This paradigm, uses optimization-based methods like SMPLify \cite{smpl} to create 3D labels for large 2D datasets \cite{coco, mpii}, which then supervise regression networks. However, existing pipelines optimize each person in isolation, leading to scene-inconsistent pGT datasets that directly limit model performance.

To improve pGT quality in multi-person scenes, various constraints have been explored. Some methods introduce constraints like a physics-based inter-penetration loss to prevent mesh intersections or a depth-ordering loss to penalize incorrect occlusions \cite{crmh, ochmr}. While effective for scenes where people are physically close or overlapping, these approaches have limited utility in common scenarios where people are spatially separated. Recent advances in monocular depth estimation, particularly powerful models like Depth Anything \cite{depthv1, depthv2}, provide rich geometric cues for scene understanding. While prior work has used depth features to estimate single-person translation \cite{blade}, our work is the first to leverage these relative depth cues within a joint optimization framework to enforce consistency among multiple people, enabling the creation of scene-consistent pGT.

Anthropometric priors offer another constraint. Prevailing works ensure realistic body shapes by regularizing shape parameters \cite{smpl, spin, smplx} or associating shape with text prompts \cite{shapesem, prompthmr}. However, these strategies are limited by their datasets, which often lack diversity in body size and hard to generalize to the full range of real-world human heights. Our approach leverage the progress in age and gender prediction \cite{mivolov1, mivolov2} and establish a statistically-informed \cite{height, growth} height prior for each individual.

% Anthropometric priors offer another constraint. Prevailing works often just regularize SMPL shape parameters \cite{smpl, spin, smplx}, which is insufficient for diverse body types and can lead to homogenous pGT. Our approach leverage the progress in age and gender prediction \cite{mivolov1, mivolov2} and establish a statistically-informed height prior for each individual.

\section{Method}
\label{sec:method}

\subsection{Preliminaries}
\paragraph{Human Model}
We adopt the unified body model from AGORA \cite{agora}, which supports varying ages by blending the SMPL \cite{smpl} and SMIL \cite{smil} templates. It is parameterized by a pose vector $\theta \in \mathbb{R}^{24 \times 3}$ and a shape vector $\beta \in \mathbb{R}^{11}$, where the final component of $\beta$ controls the age-related blending. The model function $\mathcal{M}(\theta, \beta)$ outputs a 3D mesh $V \in \mathbb{R}^{6890 \times 3}$. To place the human within the camera coordinate system, we apply a 3D translation vector $T = (x, y, z)$, resulting in the final 3D points $P = V + T$.

\paragraph{Camera Model}
We use a perspective camera model to project a 3D point $P=(X,Y,Z)$ from the camera coordinate system onto a 2D point $p $ on the image plane. We make standard simplifying assumptions that the camera has no radial distortion, and the principal point $(c_x, c_y)$ is at the image center. Following prior work \cite{camerahmr}, instead of directly estimating the camera's focal length $f$, we predict its vertical FoV $v$. The focal length can then be derived using the image height $H$: $f = \frac{H}{2 \cdot \tan(v/2)}$. The full projection function is then: $p = f \cdot \frac{(X,Y)}{Z} + (c_x, c_y)$.

\paragraph{Monocular Depth Estimation}
We leverage an off-the-shelf monocular depth estimator \cite{depthv2} to provide geometric cues. This model outputs a relative depth map $D_{\text{rel}}$, which is related to the true metric depth $D_{\text{metric}}$ by an unknown scale $s_d$ and shift $t_d$: $D_{\text{metric}} = s_d \cdot D_{\text{rel}} + t_d$. Our DTO framework is designed to solve for these global affine parameters ($s_d, t_d$) for the entire scene, thereby grounding all individuals in a consistent metric space.

\subsection{DTO Framework}
To address the scene-inconsistency of independent per-person estimations, we introduce Depth-conditioned Translation Optimization (DTO). DTO formulates the task of refining all individuals' camera-space placements as a joint optimization problem. As illustrated in Fig.~\ref{fig:dto}, our approach leverages geometric cues from monocular depth and statistical human priors within an MAP problem to find a globally optimal solution. By applying DTO to a large collection of in-the-wild images, we create DTO-Humans, a large-scale dataset for scene-consistent human reconstruction.

\subsubsection{Initial Per-Person Estimation}

We obtain initial per-person estimates via a segmentation-enhanced process \cite{harmony4d}, using Detectron2 \cite{detectron2} to generate person bounding boxes and segmentation masks. For each masked person, we employ CameraHMR \cite{camerahmr} to predict their initial mesh. Subsequently, we use 2D keypoints detected by ViTPose \cite{vitpose} to filter out inaccurate mesh estimations through a matching process (details in the supplementary material). We utilize HumanFoV \cite{camerahmr} to estimate the camera instrinsics to place human meshes in camera space.

\subsubsection{DTO Formulation and Core Components}
Our DTO framework ensures scene consistency by jointly optimizing for a global affine depth transformation, parameterized by a scale $s_d$ and a shift $t_d$. We formulate this as an MAP problem, which requires a robust prior and a way to connect it to the variables $(s_d, t_d)$ using geometric evidence. This section details the core components of our formulation.

\twocolumn[
\begin{center}
    \maketitle
    {\includegraphics[width=1\textwidth]{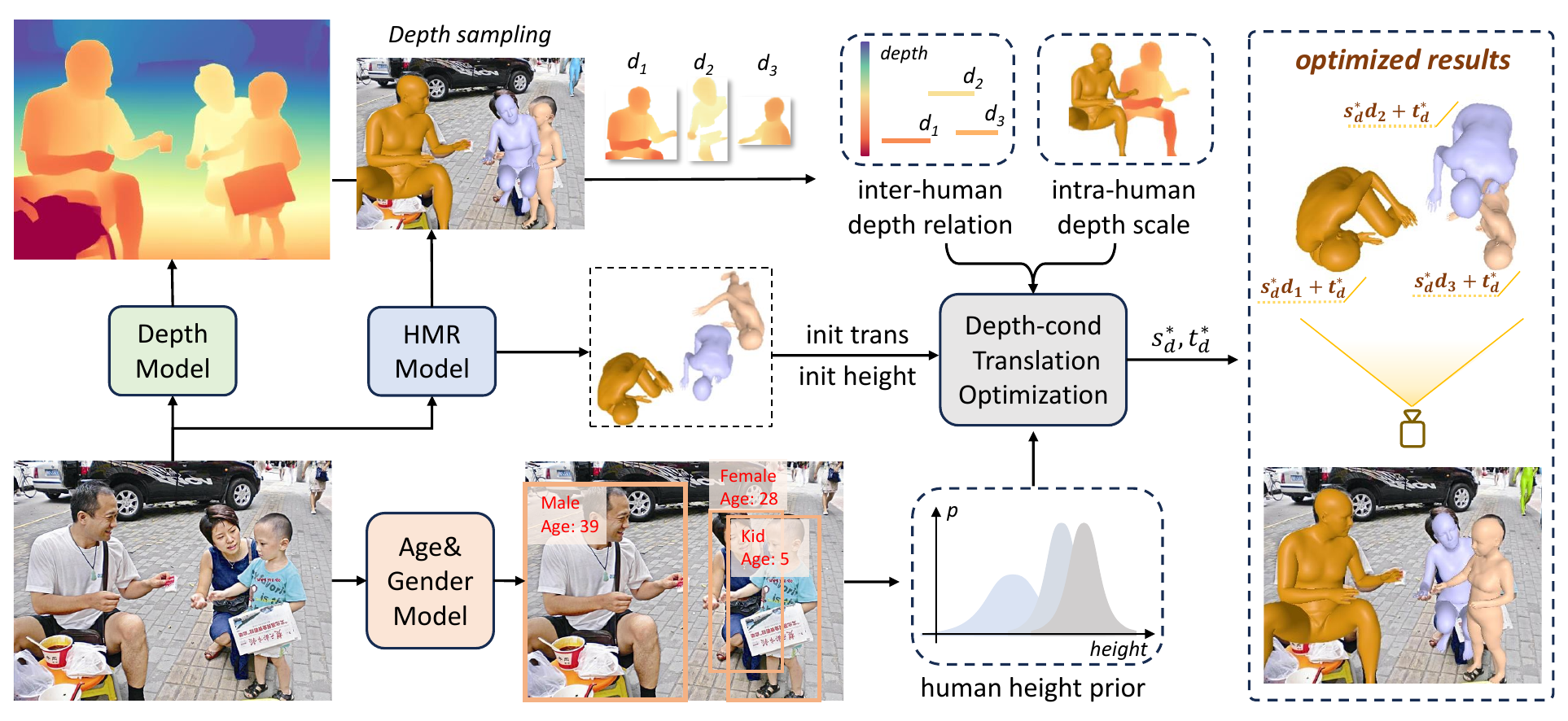}}
    \captionof{figure}{Overview of the DTO Framework. An input image is processed through three parallel streams: an HMR Model provides initial meshes; Age \& Gender Model informs a statistical height prior; and Depth Model generates a relative depth map. From the initial meshes and the depth map, we extract the inter-human depth relation and intra-human depth scale. DTO integrates these components into an MAP problem to solve for a global affine transformation, outputting a scene-consistent arrangement of all individuals.}
    \label{fig:dto}
\end{center}
]
\paragraph{Height Prior} We establish an anthropometric prior on each person's height. We utilize the expert model \cite{mivolov2} to estimate the age and gender of each individual and define the height prior for the person $i$ as a Gaussian distribution, $P_i(h_i) = \mathcal{N}(h_i | \mu_i, \sigma_i)$, where the mean $\mu_i$ and standard deviation $\sigma_i$ are derived from demographic statistics \cite{height, growth}. Details are available in the supplementary material.

\paragraph{Inter-human Depth Relation} To enforce consistent relative positioning, we extract geometric cues from a relative depth map $D_{\text{rel}}$ generated by the depth model \cite{depthv2}. For each person $i$, we define their representative depth value, $d_i$, as the mean depth sampled from $D_{\text{rel}}$ at the 2D projected locations of all their visible mesh vertices.

\paragraph{Intra-human Depth Scale} A global affine transformation ($D_{\text{metric}} = s_d D_{\text{rel}} + t_d$) without proper constraints can lead to physically implausible solutions. To prevent this, we introduce the intra-human depth scale to define a plausible range for the global scale factor $s_d$. For each person, we compute the linear regression slope $X_i$ and the correlation coefficient $w_i$ between their internal mesh depth and the values from the relative depth map. We then use the correlation coefficients as weights and compute a weighted average of these individual slopes as a scale estimate:$X = \frac{\sum_{i=1}^K w_i X_i}{\sum_{i=1}^K w_i}$. From this estimate, we establish a plausible range for $s_d$ by defining its lower and upper bounds, $X_{\min}$ and $X_{\max}$, as $X_{\min} = \alpha_1 X$ and $X_{\max} = \alpha_2 X$. The selection of hyperparameters $\alpha_1$, $\alpha_2$ is detailed in the supplementary material.

\subsubsection{DTO Objective and Solution}
With these components, we tend to find the optimal global affine transformation parameters $(s_d, t_d)$ that make the corrected height of every person most probable under their respective priors. The corrected height $h_i$ is a transformation of the initial T-pose height $\hat{h}_i$ and predicted depth $\hat{z}_i$:
\begin{equation}
h_i = \hat{h}_i \cdot \frac{s_d d_i + t_d}{\hat{z}_i}
\end{equation} This leads to the following MAP optimization problem:
\begin{equation}
\max_{s_d, t_d} \prod_{i=1}^K \mathcal{N}\left(\hat{h}_i \cdot \frac{s_d d_i + t_d}{\hat{z}_i} \bigg| \mu_i, \sigma_i\right)
\end{equation}
Taking the negative log-likelihood and incorporating our bounded intra-human scale constraint converts this into a constrained minimization problem:
\begin{equation}
\min_{s_d, t_d} \sum_{i=1}^K \frac{\left(\hat{h}_i \cdot \frac{s_d d_i + t_d}{\hat{z}_i} - \mu_i\right)^2}{\sigma_i^2} \quad \text{s.t. } X_{\min} \le s_d \le X_{\max}
\end{equation}
This objective is a convex quadratic program with analytical solution via the Karush-Kuhn-Tucker conditions. The full derivation is provided in the supplementary material.

\twocolumn[
\begin{center}
    {\includegraphics[width=0.85\textwidth]{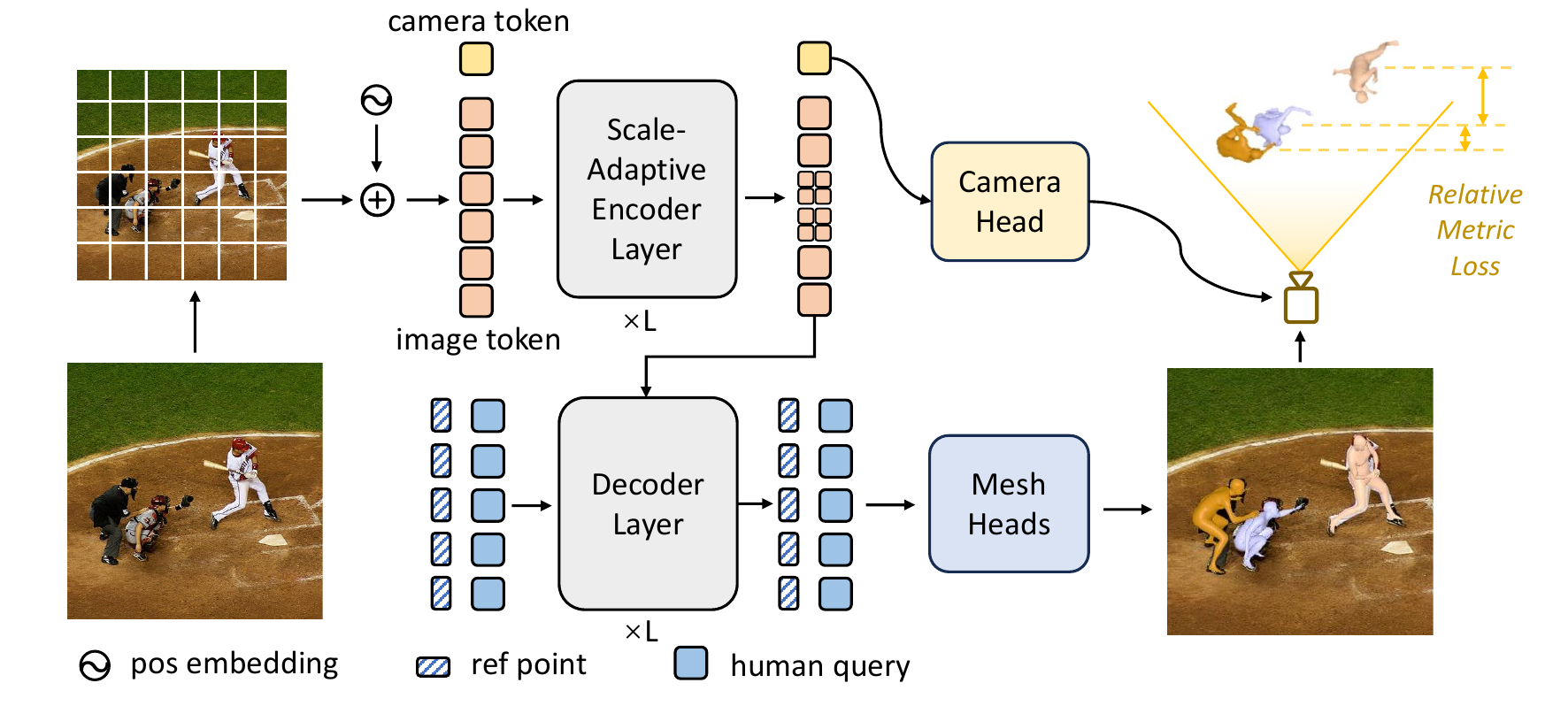}}
    \captionof{figure}{Architecture of our Metric-Aware HMR. We enhance a SAT-HMR backbone with two key innovations for true metric-scale recovery: a camera branch that predicts the camera's Field of View from global features using a dedicated camera token, and a relative metric loss that directly supervises the real-world distances between predicted individuals.}
    \label{fig:mahmr}
\end{center}
]

Once the optimal solution $(s_d^*, t_d^*)$ are found, we compute the final depth translation $z_i^* = s_d^* d_i + t_d^*$ for each person and refine their shape parameters to match the corrected heights. This process effectively grounds the entire crowd in a physically plausible metric space.

\subsubsection{DTO-Humans}
To construct a large-scale in-the-wild pGT dataset with high scene-consistency, we apply DTO to CameraHMR's \cite{camerahmr} version of 4D-Humans \cite{4dhumans} dataset, which includes 2M images from diverse in-the-wild datasets such as InstaVariety \cite{instavariety}, COCO \cite{coco}, MPII \cite{mpii}, and AI Challenger \cite{aic}. To ensure the final quality and physical plausibility of our dataset, we introduce a filtering step based on the posterior probability from our optimization. For each optimized person, we calculate their standardized residual height, i.e., $(|h_i^* - \mu_i|) / \sigma_i$. We then discard any image where the average of these residuals across all individuals exceeds a threshold of 1.5. This rigorous process filters out scenes where the optimization failed to find a solution that strongly aligns with our anthropometric priors. The resulting dataset, named \textbf{DTO-Humans}, comprises 0.56 million high-quality images and 2.7 million scene-consistent person instances, featuring dense scenes with an average of 4.8 persons per image. Further details on dataset statistics and samples are provided in the supplementary material.

\subsection{Metric-Aware HMR}
The camera-space scene-consistent pGT generated by DTO enables us to train an end-to-end network capable of real metric-scale recovery. Thus, we propose Metric-Aware HMR (MA-HMR), a novel network that directly predicts metrically accurate meshes for all subjects in the scene.

\subsubsection{Network Architecture}
Our network architecture builds upon SAT-HMR \cite{sathmr}, inheriting its Scale-Adaptive Encoder for efficient multi-scale feature extraction. As shown in Fig.~\ref{fig:mahmr}, our key architectural modification is the introduction of a dedicated camera branch. We replace the standard classification token in the ViT-style encoder, which is designed to aggregate global scene-level information, with a learnable camera token. After passing through the encoder, the output embedding of the camera token is fed into an MLP head that regresses the camera's vertical FoV. By predicting the FoV concurrently with all human individuals' mesh parameters, the network learns the intricate coupling between global scene-level human relationship and camera properties.

\subsubsection{Training Objective}

\paragraph{FoV Loss} ($\mathcal{L}_{\text{fov}}$). To supervise the camera branch, we adopt the FoV loss from HumanFoV \cite{camerahmr}, which penalizes overestimation of FoV more heavily than underestimation:
\begin{equation}
\mathcal{L}_{\text{fov}}(\upsilon_{\text{pred}}, \upsilon_{\text{gt}}) =
\begin{cases}
    3 \|\upsilon_{\text{gt}} - \upsilon_{\text{pred}}\|^2 & \text{if } \upsilon_{\text{pred}} > \upsilon_{\text{gt}} \\
    \|\upsilon_{\text{gt}} - \upsilon_{\text{pred}}\|^2 & \text{if } \upsilon_{\text{pred}} \leq \upsilon_{\text{gt}}
\end{cases}
\end{equation}

\begin{table*}[t!]
\centering
\caption{Comparison with SOTA methods on Relative Human \cite{bev} and MuPoTS \cite{mupots}.}
\label{tab:relative_human_mupots}
\small % 如果表格过宽，可以取消注释此行以缩小字体
\begin{tabular}{l c ccccc c ccc}
\toprule
% ------ 表头第一行 ------
\multirow{3}{*}{Method} & \multirow{3}{*}{f.t.} & \multicolumn{6}{c}{Relative Human} & \multicolumn{3}{c}{MuPoTS} \\
\cmidrule(lr){3-8} \cmidrule(lr){9-11}
% ------ 表头第二行 ------
& \multicolumn{6}{c}{PCDR$_{0.2}$$\uparrow$} & \multirow{2}{*}{PCK$\uparrow$} & \multirow{2}{*}{PCK(All)$\uparrow$} & \multirow{2}{*}{PCK(Match)$\uparrow$} & \multirow{2}{*}{MPJPE$\downarrow$} \\
\cmidrule(lr){3-7}
% ------ 表头第三行 ------
& & baby & kid & teen & adult & all & & & & \\
\midrule
% ------ 数据行 ------
CRMH \cite{crmh} & $\checkmark$ & 34.74 & 48.37 & 59.11 & 55.47 & 54.83 & 0.781 & / & / & / \\
ROMP \cite{romp} & $\checkmark$ & 30.08 & 48.41 & 51.12 & 55.34 & 54.81 & 0.866 & 63.0 & 67.5 & 127.8 \\
BEV \cite{bev} & $\checkmark$ & 60.77 & 67.09 & 66.07 & 69.71 & 68.27 & 0.884 & 74.7 & 78.2 & 109.0 \\
PSVT \cite{psvt} & $\checkmark$ & 64.00 & 71.29 & 70.45 & 71.95 & 71.23 & / & / & / & / \\
InstaHMR \cite{instahmr} & $\checkmark$ & \uline{66.67} & 70.85 & 72.19 & \uline{73.95} & \uline{72.86} & / & 73.7 & 76.3 & / \\
Multi-HMR \cite{multihmr} & $\checkmark$ & / & / & / & / & / & / & 84.3 & \uline{89.5}& 90.5 \\
SAT-HMR \cite{sathmr} & $\checkmark$ & 60.15 & \uline{73.95} & \textbf{74.74} & 72.52 & 72.07 & 0.908 & \textbf{87.8} & 88.8 & 92.2 \\
\midrule
CHMR \cite{camerahmr} & & 31.25 & 49.90 & 60.07 & 61.20 & 60.43 & \uline{0.921} & \uline{87.3} & 89.0 & \uline{89.5} \\
\rowcolor{gray!15}
CHMR + \textbf{DTO} & & \textbf{69.65} & \textbf{76.28} & \uline{74.53} & \textbf{74.41} & \textbf{74.16} & \textbf{0.936} & 85.0 & \textbf{90.8} & \textbf{86.3} \\
\bottomrule
\end{tabular}
\end{table*}

\paragraph{Relative Metric Loss} ($\mathcal{L}_{\text{rm}}$). To explicitly teach the network about real-world scale, we introduce a relative metric loss to enforce plausible metric relationships between individuals. This loss operates on pairs of individuals $(i, j)$ who are in close proximity, defined as those whose ground-truth relative depth distance is below threshold $\tau$ ($|z_{gt,i} - z_{gt,j}| < \tau$, with $\tau=1$m). For each qualifying pair, the loss is:
\begin{equation}
\mathcal{L}_{\text{rm}} = \log(1 + |\text{rd}_{\text{pred}} - \text{rd}_{\text{gt}}|)
\end{equation}
where $\text{rd}_{\text{gt}}$ and $\text{rd}_{\text{pred}}$ are the ground-truth and predicted relative depth distances between the individuals' root joints. The logarithmic form provides stable gradients and focuses on refining close-range predictions.

These two novel losses are integrated into a comprehensive objective function alongside the standard losses from the SAT-HMR framework \cite{sathmr}. These include the scale map loss ($\mathcal{L}_{\text{map}}$), depth loss ($\mathcal{L}_{\text{depth}}$), SMPL pose ($\mathcal{L}_{\text{pose}}$) and shape ($\mathcal{L}_{\text{shape}}$) parameter losses, a 3D joint loss ($\mathcal{L}_{\text{j3d}}$), a 2D joint reprojection loss ($\mathcal{L}_{\text{j2d}}$), L1 and GIoU bounding box losses ($\mathcal{L}_{\text{box}}$, $\mathcal{L}_{\text{giou}}$), and a focal detection loss ($\mathcal{L}_{\text{det}}$). The complete objective is a weighted sum of all components:
\begin{equation}
\mathcal{L} = \lambda_{\text{map}}\mathcal{L}_{\text{map}} + \lambda_{\text{depth}}\mathcal{L}_{\text{depth}} + \lambda_{\text{pose}}\mathcal{L}_{\text{pose}} + \lambda_{\text{shape}}\mathcal{L}_{\text{shape}}
\end{equation}
$$
\quad + \lambda_{\text{j3d}}\mathcal{L}_{\text{j3d}} + \lambda_{\text{j2d}}\mathcal{L}_{\text{j2d}} + \lambda_{\text{box}}\mathcal{L}_{\text{box}} + \lambda_{\text{giou}}\mathcal{L}_{\text{giou}}\\ 
$$
$$
+\lambda_{\text{det}}\mathcal{L}_{\text{det}} + \lambda_{\text{fov}}\mathcal{L}_{\text{fov}} + \lambda_{\text{rm}}\mathcal{L}_{\text{rm}}
$$
% \begin{equation}
% \mathcal{L} = \lambda_{\text{map}}\mathcal{L}_{\text{map}} + \lambda_{\text{depth}}\mathcal{L}_{\text{depth}} + \lambda_{\text{pose}}\mathcal{L}_{\text{pose}} + \lambda_{\text{shape}}\mathcal{L}_{\text{shape}} \\
% + \lambda_{\text{j3d}}\mathcal{L}_{\text{j3d}} + \lambda_{\text{j2d}}\mathcal{L}_{\text{j2d}} + \lambda_{\text{box}}\mathcal{L}_{\text{box}} + \lambda_{\text{giou}}\mathcal{L}_{\text{giou}}\\ + \lambda_{\text{det}}\mathcal{L}_{\text{det}} 
% + \lambda_{\text{fov}}\mathcal{L}_{\text{fov}} + \lambda_{\text{rm}}\mathcal{L}_{\text{rm}}
% \end{equation}
where the $\lambda$ terms are the corresponding loss weights. 

% This comprehensive objective ensures that our model not only recovers accurate poses and shapes but also places them correctly within a metrically consistent 3D scene.

\section{Experiments}
\label{sec:experiments}

\subsection{Implementation Details}
For DTO, We set Detectron2's \cite{detectron2} detection threshold at 0.5 and nms threshold at 0.7. We discard humans whose visible keypoints predicted by ViTPose-Base \cite{vitpose} are less than five. We use Depth-Anything-v2-Large \cite{depthv2} for predicting the relative depth map and Mivolo-v2 \cite{mivolov2} to predict human age and gender based on their body and face crops. 

For MA-HMR, we adopt ViT-Base \cite{vit, dinov2} as the backbone. We initialize our model with the SAT-HMR's \cite{sathmr} stage1 checkpoint pretrained on AGORA \cite{agora}, BEDLAM \cite{bedlam}, COCO \cite{coco}, MPII \cite{mpii}, CrowdPose \cite{crowdpose}, and Human3.6M \cite{h36m}, and continue training on AGORA, BEDLAM and CameraHMR's \cite{camerahmr} version of 4D-Humans dataset \cite{instavariety, coco, mpii, aic} for 5 epochs with denoising strategy \cite{dn, sathmr}. In the second stage, we add camera branch and continue training for 5 epochs with FoV loss. For pGTs from 4D-Humans, we follow SAT-HMR stage1, only to supervise the projected 2D keypoints. In the third stage, we train the model on AGORA, BEDLAM and DTO-Humans for 5 epochs with full loss. In the final stage, we finetune the model for each benchmark following \cite{sathmr, multihmr, bev}. For all training stages, we use the AdamW optimizer \cite{adamw} with a learning rate of 1e-5 and weight decay of 1e-4. Data augmentation techniques include random rotation, horizontal flipping, scaling, and cropping. Training is conducted on two NVIDIA A800 GPUs with a total batch size of 64. 

During evaluation of MA-HMR, we filter out low-confidence detections using a threshold of 0.3. Evaluation metrics on 3DPW \cite{3dpw}, Hi4D \cite{hi4d}, CMU Panoptic \cite{panoptic} and MuPoTS \cite{mupots} include Mean Per-Joint Position Error (MPJPE), Procrustes Aligned MPJPE (PA-MPJPE), and Per-Vertex Error (PVE), all reported in millimeters (mm). The Percentage of Correct Depth Relations (PCDR$_{0.2}$) and the Percentage of Correctly estimated 2D Keypoints (PCK) using a threshold of 1/7 of human's height are evaluated on Relative Human following \cite{bev}. The Percentage of Correctly estimated 3D Keypoints (PCK) using a threshold of 15 cm is evaluated on MuPoTS. Height error (mm) is evaluated on 3DPW and Hi4D by comparing the T-pose height from the predicted shape parameters to the ground-truth.

\subsection{Quantitative Comparison of DTO}
To quantify the contribution of our DTO framework, we conduct an experiment to apply it as a post-processing refinement step to a state-of-the-art single-person HMR model CameraHMR \cite{camerahmr} (CHMR). As shown in Table~\ref{tab:relative_human_mupots}, we compare the output of CHMR and the output after refinement by our DTO framework with SOTA methods. It is important to note that neither of CHMR and DTO is fine-tuned (f.t.) on the benchmark-specific training sets, allowing for the evaluation of DTO's generalization.

\begin{table*}[t!]
\centering
\setlength{\tabcolsep}{5pt}
\caption{Comparison with SOTA methods on 3DPW \cite{3dpw}, CMU Panoptic \cite{panoptic} and MuPoTS \cite{mupots}.}
\label{tab:performance_comparison_3dpw}
\small % 使用稍小的字体以适应表格宽度
\begin{tabular}{l c ccc ccccc cc}
\toprule
% ------ 表头第一行 ------
\multirow{2}{*}{Method} & \multirow{2}{*}{f.t.} & \multicolumn{3}{c}{3DPW} & \multicolumn{5}{c}{CMU Panoptic (MPJPE$\downarrow$)} & \multicolumn{2}{c}{MuPoTS (PCK$\uparrow$)} \\
% ------ 表头分割线 ------
\cmidrule(lr){3-5} \cmidrule(lr){6-10} \cmidrule(lr){11-12}
% ------ 表头第二行 ------
& & MPJPE$\downarrow$ & PA-MPJPE$\downarrow$ & MVE$\downarrow$ & haggling & mafia & ultimatum & pizza & mean & All & Match \\
\midrule
% ------ 数据部分 - 第一组 ------
CMRH \cite{crmh} & $\checkmark$ & / & / & / & 129.6 & 133.5 & 153.0 & 156.7 & 143.2 & / & / \\
3DCrowdNet \cite{3dcrowdnet} & $\checkmark$ & 81.7 & 51.5 & 98.3 & 109.6 & 135.9 & 129.8 & 135.6 & 127.6 & 72.7 & 73.3 \\
ROMP \cite{romp} & $\checkmark$ & 76.6 & 47.3 & 93.4 & 110.8 & 135.9 & 129.8 & 135.6 & 127.6 & 63.0 & 67.5 \\
BEV \cite{bev} & $\checkmark$ & 78.5 & 46.9 & 92.3 & 90.7 & 103.7 & 113.1 & 125.2 & 109.5 & 74.7 & 78.2 \\
PSVT \cite{psvt} & $\checkmark$ & 75.5 & 45.7 & 84.9 & 88.7 & 97.9 & 115.2 & 121.1 & 105.7 & / & / \\
Insta-HMR \cite{instahmr} & $\checkmark$ & 77.7 & 46.3 & 87.0 & 84.8 & 99.8 & 102.6 & 121.7 & 104.5 & 73.7 & 76.3 \\
Multi-HMR \cite{multihmr} & $\checkmark$ & 61.4 & 41.7 & 75.9 & / & / & / & / & 96.5 & 84.3 & 89.5 \\
SAT-HMR \cite{sathmr} & $\checkmark$ & 63.6 & 41.6 & 73.7 & 67.9 & 78.5 & 95.8 & 94.6 & 84.2 & 87.8 & 88.8 \\
\midrule
% ------ 数据部分 - 第二组 (消融实验) ------
SAT-HMR stage1 & & 81.0 & 52.7 & 94.5 & 101.2 & 120.5 & 135.4 & 120.6 & 119.3 & 83.2 & 86.6 \\
SAT-HMR w. 4D & & 68.8 & 44.5 & 80.5 & 80.5 & 102.2 & 113.6 & 96.7 & 98.5 & 85.4 & 89.0 \\
% \rowcolor{gray!10}
\textbf{MA-HMR} w. 4D & & 65.4 & 42.6 & 76.9 & 79.0 & 98.6 & 114.8 & 98.6 & 97.5 & 83.0 & 89.5 \\
% \rowcolor{gray!10}
SAT-HMR \textbf{w. DH} & & \uline{63.6} & \uline{40.5} & \uline{74.4} & \textbf{64.3} & \uline{78.6} &\uline{87.7} & \textbf{85.5} & \textbf{79.6} & \uline{86.9} & \textbf{90.0} \\
\rowcolor{gray!15}
\textbf{MA-HMR w. DH} & & \textbf{62.9} & \textbf{40.1} & \textbf{73.5} & \uline{66.2} & \textbf{76.7} & \textbf{87.0} & \uline{87.4} & \textbf{79.6} & \textbf{87.0} & \uline{89.8} \\
\midrule
% ------ 数据部分 - 第三组 (精调后) ------
% \rowcolor{gray!10}
SAT-HMR \textbf{w. DH} & $\checkmark$ & \uline{60.1} & \uline{37.7} & \uline{69.7} & \uline{64.5} & \uline{72.7} & \uline{88.7} & \textbf{83.6} & \uline{76.9} & / & / \\
\rowcolor{gray!15}
\textbf{MA-HMR w. DH} & $\checkmark$ & \textbf{58.5} & \textbf{36.3} & \textbf{67.9} & \textbf{62.3} & \textbf{72.3} & \textbf{84.3} & \uline{85.4} & \textbf{76.3} & / & / \\
\bottomrule
\end{tabular}
\end{table*}

On the Relative Human dataset, which is challenging due to its diverse ages and complex relative depths, the improvement of DTO is remarkable. The baseline CHMR, processing each person in isolation and neglecting the modeling of kids, achieves a PCDR$_{0.2}$ (all) of only 60.43. By simply applying our DTO to the baseline, the score dramatically increases by 13.73 points to 74.16, surpassing all fine-tuned SOTA methods, including the previous leader InstaHMR (+1.3). The segmentation-enhanced mesh initialization also contributes to a top-performing PCK score of 0.936, indicating superior 2D alignment.

The evaluation on the MuPoTS dataset further validates our approach. Our CHMR + DTO pipeline achieves lowest MPJPE among all competing methods. This result proves that even in less crowded scenes (with 2 or 3 adults), DTO can also solve for a consistent scale that results in more plausible human heights and thus leads to superior 3D human joints accuracy. These experiments confirm that DTO is a powerful and robust framework for enhancing scene-level consistency and metric accuracy.

\subsection{Quantitative Comparison of MA-HMR}

We first evaluate our model's understanding of scene-level relative relationships on the Relative Human dataset. As shown in Table~\ref{tab:comparison_rh}, our fine-tuned MA-HMR sets a new SOTA with a PCDR$_{0.2}$ (all) of 75.35, surpassing both the SAT-HMR baseline (72.89) and our optimization-based CHMR + DTO (74.16). This demonstrates the advantage of end-to-end fine-tuning, allowing the model to specialize in the dataset's unique age distributions and complex relative depth cues. Notably, even without fine-tuning, our MA-HMR trained on DTO-Humans (72.90) is already on par with the fine-tuned SAT-HMR (72.89). This highlights the strong generalization provided by our model architecture when trained on our high-quality dataset.

\begin{table}[t]
\centering
\setlength{\tabcolsep}{2.5pt}
\small % 使用稍小的字体以适应表格宽度
% \footnotesize % 使用稍小的字体以适应表格宽度
\caption{Results comparison on Relative Human \cite{bev}.}
\label{tab:comparison_rh}
\begin{tabular}{l c ccccc c}
\toprule
\multirow{2}{*}{Method} & \multirow{2}{*}{f.t.} & \multicolumn{5}{c}{PCDR$_{0.2}$$\uparrow$} & \multirow{2}{*}{PCK$\uparrow$} \\
\cmidrule(lr){3-7}
 & & baby & kid & teen & adult & all & \\
% \midrule
% InstaHMR & $\checkmark$ & \uline{66.67} & 70.85 & 72.19 & \uline{73.95} & 72.86 & / \\
\midrule
% \rowcolor{gray!10}
CHMR + \textbf{DTO} & & \textbf{69.65} & \textbf{76.28} & 74.53 & \uline{74.41} & \uline{74.16} & \textbf{0.936} \\
% \rowcolor{gray!10}
SAT-HMR \textbf{w. DH} & & 59.14 & 68.98 & 66.20 & 68.83 & 68.30 & \uline{0.927} \\
\rowcolor{gray!15}
\textbf{MA-HMR w. DH} & & 58.64 & 72.35 & \textbf{76.83} & 73.43 & 72.90 & 0.925 \\
\midrule
% \rowcolor{gray!10}
SAT-HMR \textbf{w. DH} & $\checkmark$ & 65.56 & 74.24 & 73.85 & 73.31 & 72.89 & 0.924 \\
\rowcolor{gray!15}
\textbf{MA-HMR w. DH} & $\checkmark$ & \uline{67.76} & \uline{75.57} & \uline{76.81} & \textbf{75.77} & \textbf{75.35} & 0.926 \\
\bottomrule
\end{tabular}
\end{table}

We then evaluate human pose and shape accuracy on the classic benchmarks 3DPW, CMU Panoptic and MuPoTS with results in Table~\ref{tab:performance_comparison_3dpw}. The superior generalization of our dataset is evident: without benchmark-specific fine-tuning, SAT-HMR \cite{sathmr} trained on DTO-Humans (w. DH) significantly outperforms the same model trained on CameraHMR's version of 4D-Humans (w. 4D). This is especially evident on CMU Panoptic, where the mean MPJPE drops from 98.5 mm to 79.6 mm. Our MA-HMR architecture (MA-HMR w. DH) further improves upon this, achieving 62.9 mm on 3DPW. After fine-tuning, MA-HMR sets a new state-of-the-art on 3DPW, achieving 58.5 mm MPJPE and 36.3 mm PA-MPJPE. It also records the lowest mean error on CMU Panoptic at 76.3 mm.

\begin{table}[t!]
\centering
\small % 使用稍小的字体以适应表格宽度
\caption{Comparison with SOTA methods on Hi4D \cite{hi4d}.}
\label{tab:hi4d_comparison}
\begin{tabular}{l c ccc}
\toprule
% ------ 表头第一行 ------
Method & f.t. & MPJPE$\downarrow$ & PA-MPJPE$\downarrow$ & MVE$\downarrow$ \\
\midrule
% ------ 数据部分 - 第一组 (与现有方法比较) ------
CLIFF \cite{cliff}      & $\checkmark$ & 91.3 & 53.6 & 109.6 \\
4D-Humans \cite{4dhumans}      & $\checkmark$ & 72.1 & 52.4 & 88.6 \\
BEV \cite{bev}      & $\checkmark$ & 91.8 & 52.2 & 101.2 \\
TRACE \cite{trace}         & $\checkmark$ & 83.8 & 60.4 & / \\
GroupRec \cite{grouprec}      & $\checkmark$ & 82.4 & 51.6 & 88.6 \\
BUDDI \cite{buddi}         & $\checkmark$ & 96.8 & 70.6 & 116.0 \\
CloseInt \cite{closeint}      & $\checkmark$ & 63.1 & 47.5 & 76.4 \\
Fang et al. \cite{fang2024capturing}   & $\checkmark$ & 75.0 & 59.7 & / \\
ReconClose \cite{reconclose}    & $\checkmark$ & 59.1 & 44.3 & 72.0 \\
% SAT-HMR \cite{sathmr}       & $\checkmark$ & 51.1 & 38.2 & 62.5 \\
\midrule
% ------ 数据部分 - 第二组 (消融实验) ------
SAT-HMR stage1 &              & 95.5 & 62.0 & 111.7 \\
SAT-HMR w. 4D  &              & 74.0 & 50.0 & 90.5 \\
% \rowcolor{gray!10}
\textbf{MA-HMR} w. 4D   &              & 68.3 & 47.3 & 85.1 \\
% \rowcolor{gray!10}
SAT-HMR \textbf{w. DH}  &              & \uline{61.0} & \uline{46.4} & \uline{76.9} \\
\rowcolor{gray!15}
\textbf{MA-HMR w. DH}   &              & \textbf{60.2} & \textbf{45.6} & \textbf{76.0} \\
\midrule
% ------ 数据部分 - 第三组 (最终精调结果) ------
% \rowcolor{gray!10}
SAT-HMR \textbf{w. DH}  & $\checkmark$ & \uline{44.3} & \uline{34.3} & \uline{54.9} \\
\rowcolor{gray!15}
\textbf{MA-HMR w. DH}   & $\checkmark$ & \textbf{43.5} & \textbf{33.8} & \textbf{54.2} \\
\bottomrule
\end{tabular}
\end{table}

On the Hi4D dataset, which focuses on close human-human interaction, our method again shows superior performance (Table~\ref{tab:hi4d_comparison}). Simply training the baseline model on DTO-Humans (SAT-HMR w. DH) improves MPJPE from 74.0 mm (SAT-HMR w. 4D) to 61.0 mm. After fine-tuning, our MA-HMR w. DH model reaches 43.5 mm MPJPE and 33.8 mm PA-MPJPE, outperforming all previous methods.

We validate the metric accuracy on human scale which contributes to the per-joint accuracy (Table~\ref{tab:height_error}). The reduction in height error when switching from w. 4D to w. DH (models here are not benchmark-specific finetuned) confirms that our dataset provides a more reliable height supervision. Furthermore, MA-HMR achieves the lowest height error, demonstrating that the relative metric loss effectively guides the network to learn plausible human scales. 

\begin{table}[t!]
\centering
\setlength{\tabcolsep}{12pt}
\caption{Comparison of height error (mm)$\downarrow$ on 3DPW and Hi4D.}
\label{tab:height_error}
\small % 鉴于表格较宽，使用 \small 字体
\begin{tabular}{l c c}
\toprule
% ------ 表头第一行 ------
Method & 3DPW & Hi4D \\
\midrule
% ------ 数据行 ------
SAT-HMR w. 4D & 49.0 & 111.1 \\
MA-HMR w. 4D & 41.5 & 86.9 \\
SAT-HMR w. DH & 38.2 & 38.3 \\
\rowcolor{gray!15}
MA-HMR w. DH & \textbf{36.9} & \textbf{35.0} \\
\bottomrule
\end{tabular}
\end{table}

\subsection{Ablation Study}
\paragraph{Ablation on DTO Components}
As detailed in Table~\ref{tab:ablation_dto}, we evaluate the contribution of each component in our DTO framework on the Relative Human dataset, starting from the baseline CHMR model. By incorporating segmentation enhanced initial per-person estimation (+S), we slightly improve 2D pose estimation (PCK 0.921 $\rightarrow$ 0.936) as the model can better focus on the target human. By leveraging Inter-human Depth relation for translation optimization (+D), the PCDR$_{0.2}$ (all) score jumps from 60.89 to 72.15. This demonstrates that optimizing for relative depth is the most critical factor in resolving scene-wide layout. Finally, we add the intra-human depth scale constraint (+X). This further improves the PCDR$_{0.2}$ (all) score from 72.15 to 74.16. This shows that enforcing this scale constraint prevents the optimization from settling on physically unrealistic scale factors and achieve a more robust estimation.

\begin{table}[t!]
\centering
\setlength{\tabcolsep}{4pt}
\caption{Ablation studies on DTO on Relative Human. S: segmentation-enhanced inference. D: optimization based on inter-human depth relation. X: intra-human depth scale constraints.}
\label{tab:ablation_dto}
\small % 鉴于表格较宽，使用 \small 字体
\begin{tabular}{l ccccc c}
\toprule
% ------ 表头第一行 ------
\multirow{2}{*}{Method} & \multicolumn{5}{c}{PCDR$_{0.2}$$\uparrow$} & \multirow{2}{*}{PCK$\uparrow$} \\
% ------ 表头横线 ------
\cmidrule(lr){2-6}
% ------ 表头第二行 ------
 & baby & kid & teen & adult & all & \\
\midrule
% ------ 数据行 ------
CHMR & 31.25 & 49.90 & 60.07 & 61.20 & 60.43 & 0.921 \\
CHMR+S & 30.80 & 50.93 & 62.41 & 61.29 & 60.89 & 0.936 \\
CHMR+S+D & 65.77 & 72.40 & 74.53 & 72.38 & 72.15 & 0.936 \\
CHMR+S+D+X & \textbf{69.65} & \textbf{76.28} & \textbf{74.53} & \textbf{74.41} & \textbf{74.16} & \textbf{0.936} \\
\bottomrule
\end{tabular}
\end{table}

\begin{table}[t!]
\centering
\caption{Ablation studies on MA-HMR.}
\label{tab:ablation_mahmr}
% 1. 使用 \small 字体
\small
% 2. 减小列间距 (默认是 6pt)
\setlength{\tabcolsep}{6pt}
% 3. 使用 @{} 移除表格两侧的空白
\begin{tabular}{@{}cc cc cc@{}}
\toprule
% ------ 表头第一行 ------
\multirow{2}{*}{camera} & \multirow{2}{*}{$\mathcal{L}_{\text{rm}}$} & \multicolumn{2}{c}{3DPW} & \multicolumn{2}{c}{Hi4D} \\
% ------ 表头横线 ------
\cmidrule(lr){3-4} \cmidrule(lr){5-6}
% ------ 表头第二行 ------
 & & MPJPE$\downarrow$ & MVE$\downarrow$ & MPJPE$\downarrow$ & MVE$\downarrow$ \\
\midrule
% ------ 数据行 ------
 & &  63.6 & 74.4 & 61.0 & 76.9 \\
$\checkmark$ &  & 63.1 & 74 & 61.7 & 77.4 \\
$\checkmark$ & $\checkmark$ & \textbf{62.9}  & \textbf{73.5} & \textbf{60.2}  & \textbf{76.0} \\
\bottomrule
\end{tabular}
\end{table}

\paragraph{Ablation on MA-HMR Architecture}In Table~\ref{tab:ablation_mahmr}, we analyze our architectural choices for MA-HMR. All models are trained on DTO-Humans without benchmark-specific finetuning. We find that adding only the camera branch improves performance on 3DPW but slightly degrades on Hi4D (MPJPE 61.0 $\rightarrow$ 61.7). We attribute this to Hi4D's uniform camera FoV settings ($59.53^\circ \pm 0.07^\circ$), which align with the baseline's fixed $60^\circ$ default. The unconstrained FoV prediction introduces unnecessary variance. However, combining the camera branch with our relative metric loss ($\mathcal{L}_{\text{rm}}$) yields the best performance across all benchmarks. This confirms that $\mathcal{L}_{\text{rm}}$ provides essential supervision, enabling the network to effectively resolve the ambiguity between camera intrinsics and human scale.

\section{Conclusion}
\label{sec:conclusion}
In this paper, we introduce DTO, an optimization framework for scene-consistent pGT generation, resulting in the DTO-Humans dataset. We also propose MA-HMR, an end-to-end network featuring a camera branch and a relative metric loss to achieve metric-scale recovery. Extensive experiments demonstrate that our method achieves SOTA performance in human relative depth reasoning and pose and shape accuracy, confirming that our DTO-Humans dataset and MA-HMR effectively bridge the gap between relative scene understanding and absolute metric reconstruction.

{
    \small
    \bibliographystyle{ieeenat_fullname}
    \normalem
    \bibliography{main}
}

\clearpage
\setcounter{page}{1}
\maketitlesupplementary

% \section{Rationale}
% \label{sec:rationale}
% % 
% Having the supplementary compiled together with the main paper means that:
% % 
% \begin{itemize}
% \item The supplementary can back-reference sections of the main paper, for example, we can refer to \cref{sec:intro};
% \item The main paper can forward reference sub-sections within the supplementary explicitly (e.g. referring to a particular experiment); 
% \item When submitted to arXiv, the supplementary will already included at the end of the paper.
% \end{itemize}
% % 
% To split the supplementary pages from the main paper, you can use \href{https://support.apple.com/en-ca/guide/preview/prvw11793/mac#:~:text=Delete%20a%20page%20from%20a,or%20choose%20Edit%20%3E%20Delete).}{Preview (on macOS)}, \href{https://www.adobe.com/acrobat/how-to/delete-pages-from-pdf.html#:~:text=Choose%20%E2%80%9CTools%E2%80%9D%20%3E%20%E2%80%9COrganize,or%20pages%20from%20the%20file.}{Adobe Acrobat} (on all OSs), as well as \href{https://superuser.com/questions/517986/is-it-possible-to-delete-some-pages-of-a-pdf-document}{command line tools}.

\section{DTO}
\subsection{Analytical Solution to the DTO Problem}
We aim to solve the following constrained quadratic minimization problem, where the scale parameter $s_d$ is bounded by a plausible range derived from the intra-human depth scale:
\begin{equation}
\min_{s_d, t_d} \sum_{i=1}^K \frac{\left(\hat{h}_i \cdot \frac{s_d d_i + t_d}{\hat{z}i} - \mu_i\right)^2}{\sigma_i^2} \quad \text{s.t.} \quad X_{\min} \le s_d \le X_{\max}
\end{equation}
\paragraph{Simplification of Terms}
To simplify the derivation, we define $a_i = \frac{\hat{h}_i d_i}{\hat{z}_i \sigma_i}$, $b_i = \frac{\hat{h}_i}{\hat{z}_i \sigma_i}$, and $c_i = \frac{\mu_i}{\sigma_i}$.
The objective function, which we denote as $L(s_d, t_d)$, can now be rewritten as a simple sum of squares:
\begin{equation}
L(s_d, t_d) = \sum_{i=1}^K (a_i s_d + b_i t_d - c_i)^2
\end{equation}
The problem has two inequality constraints, which we write in the standard form $g(x) \le 0$:
\begin{align}
g_1(s_d, t_d) &= X_{\min} - s_d \leq 0 \\
g_2(s_d, t_d) &= s_d - X_{\max} \leq 0
\end{align}
\paragraph{The Lagrangian and KKT Conditions}
The Lagrangian $\mathcal{L}_{\text{g}}$ for this problem involves two Lagrange multipliers, $\lambda_1$ and $\lambda_2$, corresponding to the two constraints:
\begin{equation}
\mathcal{L}_{\text{g}}(s_d, t_d, \lambda_1, \lambda_2) = L(s_d, t_d) + \lambda_1(X_{\min} - s_d) + \lambda_2(s_d - X_{\max})
\end{equation}
The KKT conditions for optimality are:
\begin{subequations}
\label{eq:KKT_conditions_bounded} % 为整个方程组添加一个标签
\begin{empheq}[left=\empheqlbrace]{flalign}
\frac{\partial \mathcal{L}_{\text{g}}}{\partial s_d} =\sum_{i=1}^K 2a_i (a_i s_d + b_i t_d - c_i) - \lambda_1 + \lambda_2 = 0, \label{eq:kkt_bounded_1} \\
\frac{\partial \mathcal{L}_{\text{g}}}{\partial t_d} =\sum_{i=1}^K 2b_i (a_i s_d + b_i t_d - c_i) = 0, \label{eq:kkt_bounded_2} \\
X_{\min} - s_d \leq 0, \label{eq:kkt_bounded_3} \\
s_d - X_{\max} \leq 0, \label{eq:kkt_bounded_4} \\
\lambda_1, \lambda_2 \geq 0, \label{eq:kkt_bounded_5} \\
\lambda_1 (X_{\min} - s_d) = 0, \label{eq:kkt_bounded_6} \\
\lambda_2 (s_d - X_{\max}) = 0 \label{eq:kkt_bounded_7}
\end{empheq}
\end{subequations}
\paragraph{Solving the System}
The nature of the convex quadratic objective and the linear constraints allows for a straightforward solution strategy. The approach is to first solve the unconstrained minimum of the objective function, assuming both constraints are inactive ($\lambda_1 = \lambda_2 = 0$). In this scenario, the stationarity conditions (Eqs. \eqref{eq:kkt_bounded_1} and \eqref{eq:kkt_bounded_2}) simplify to a standard 2x2 linear system:
\begin{subequations}
\label{eq:linear_system_bounded}
\begin{empheq}[left=\empheqlbrace]{align}
\left( \sum a_i^2 \right) s_d + \left( \sum a_i b_i \right) t_d &= \sum a_i c_i \\
\left( \sum a_i b_i \right) s_d + \left( \sum b_i^2 \right) t_d &= \sum b_i c_i
\end{empheq}
\end{subequations}
Solving this system yields the unconstrained solution, denoted $(s_{d, \text{unc}}, t_{d, \text{unc}})$.
We now check where $s_{d, \text{unc}}$ falls relative to the interval $[X_{\min}, X_{\max}]$. There are three possible outcomes:

\textbf{Case 1: The unconstrained solution is feasible.}
If $X_{\min} \le s_{d, \text{unc}} \le X_{\max}$, the unconstrained minimum already satisfies the constraints. In this case, the KKT conditions with $\lambda_1 = \lambda_2 = 0$ are fully satisfied. The global optimum is the unconstrained solution:
\begin{equation}
(s_d^*, t_d^*) = (s_{d, \text{unc}}, t_{d, \text{unc}})
\end{equation}

\textbf{Case 2: The unconstrained solution is below the lower bound.}
If $s_{d, \text{unc}} < X_{\min}$, the convexity of the objective function implies that the minimum over the feasible set must lie on the boundary closest to the unconstrained minimum. Therefore, the optimal scale $s_d^*$ is clamped to the lower bound:
\begin{equation}
s_d^* = X_{\min}
\end{equation}
This corresponds to the KKT case where the lower bound is active ($\lambda_1 > 0$) and the upper bound is inactive ($\lambda_2 = 0$). We substitute $s_d^* = X_{\min}$ into the second stationarity condition (Eq. \eqref{eq:kkt_bounded_2}), which is independent of the multipliers, to solve for the optimal $t_d^*$:
\begin{equation}
\left( \sum b_i^2 \right) t_d^* = \sum (b_i c_i - a_i b_i X_{\min})
\end{equation}
\begin{equation}
t_d^* = \frac{\sum b_i c_i - X_{\min} \sum a_i b_i}{\sum b_i^2}
\end{equation}

\textbf{Case 3: The unconstrained solution is above the upper bound.}
If $s_{d, \text{unc}} > X_{\max}$, by the same logic, the solution is clamped to the upper bound:
\begin{equation}
s_d^* = X_{\max}
\end{equation}
This corresponds to the KKT case where the upper bound is active ($\lambda_2 > 0$) and the lower bound is inactive ($\lambda_1 = 0$). We substitute $s_d^* = X_{\max}$ into Eq. \eqref{eq:kkt_bounded_2} to find the corresponding $t_d^*$:
\begin{equation}
t_d^* = \frac{\sum b_i c_i - X_{\max} \sum a_i b_i}{\sum b_i^2}
\end{equation}

This procedure of solving the unconstrained problem and then clamping the solution to the feasible interval [$X_{\min},X_{\max}$] is guaranteed to find the unique global minimum of this convex, constrained quadratic program.

\begin{figure}[htbp]
    \includegraphics[width=\linewidth]{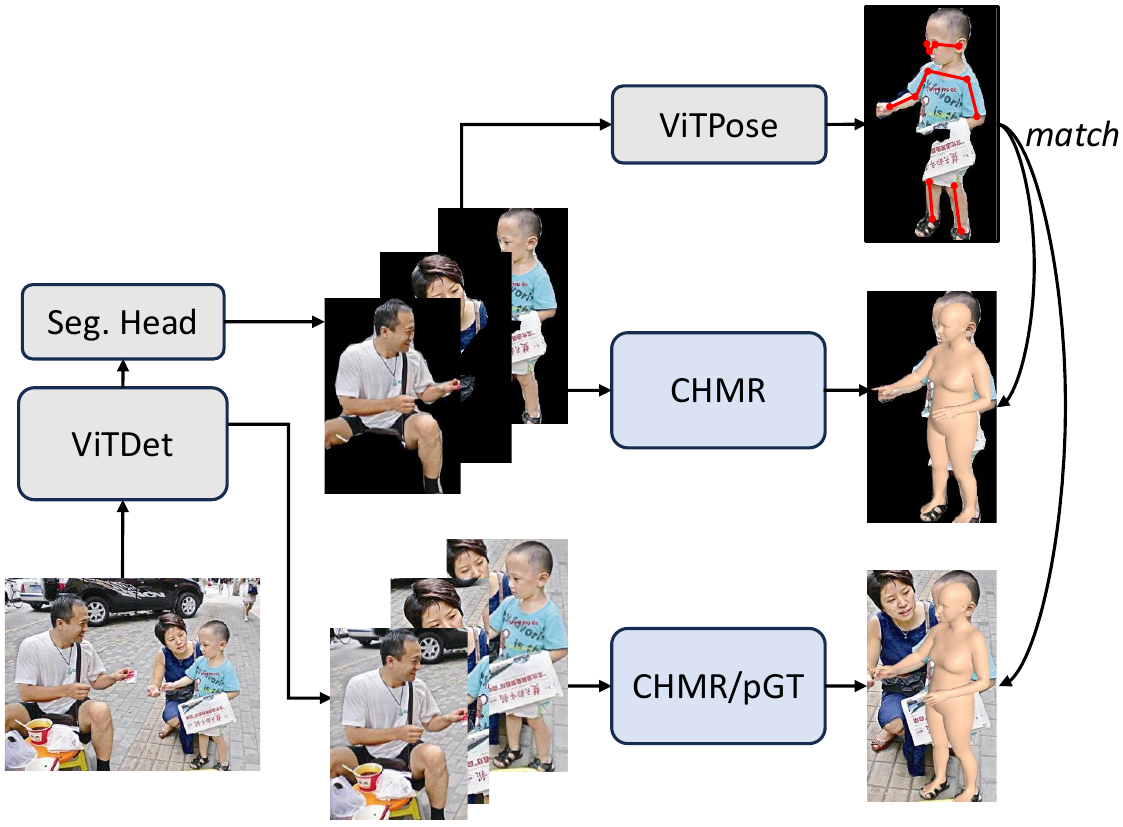}
    \caption{Initial Per-Person Estimation Pipeline.}
    \label{fig:hmr_init}
\end{figure}

\subsection{Initial Per-Person Estimation}
To leverage 2D landmarks to filter out inaccurate mesh estimations, we introduce a two-stage matching process. The 2D keypoints detected by ViTPose on the masked crops serve as the \textit{query} set for this process. Specifically, an instance is considered a valid match if over half of its visible keypoints align with projected joints. An individual keypoint-to-joint correspondence is deemed successful if their pixel distance is less than half the person's head height.

In the first stage, we match these query keypoints against a high-fidelity target. For datasets with available pGT, we use the keypoints projected from the pGT mesh as the initial target. If pGT is not available, we use the keypoints projected from an initial mesh predicted by CameraHMR on the original, unmasked crop to leverage full crop context.

In the second stage, any query keypoints that remain unmatched are subsequently compared against the keypoints projected from the mesh inferred from the masked crop. This allows us to find correspondences for bodies that might be better aligned in the segmentation-enhanced prediction.

This hierarchical strategy ensures we prioritize high-quality correspondences while still leveraging the fine-grained alignment from the masked prediction. 

\subsection{Intra-human Scale Factor Bounds}
The hyperparameters $\alpha_1$ and $\alpha_2$, which define the lower and upper bounds for the global depth scale $s_d$, are set dynamically based on the scene's spatial layout. We first identify quasi-planar scenes by checking if the variance of inter-person depths is smaller than the average intra-person depth variance. In such cases, where all individuals are roughly equidistant from the camera, the intra-human scale X is the most reliable cue. Therefore, we set $\alpha_1$ = $\alpha_2$ = 1, effectively fixing the scale factor to $s_d = X$ and solving for $t_d$ only.

Otherwise, for scenes with clear depth separation, we set $\alpha_1$ = 1 and $\alpha_2$ = 5. This wider range is motivated by the behavior of initial camera estimation pipeline (HumanFoV). Their FoV loss leads to larger predicted focal lengths to prevent artifacts, which in turn results in overestimated camera-space human depth translations ($z_i$). By allowing $s_d$ to be larger than the intra-human scale X, we give the optimizer a sufficient range to find a globally consistent scale.

\subsection{Height Priors}
\begin{figure}
    \includegraphics[width=\linewidth]{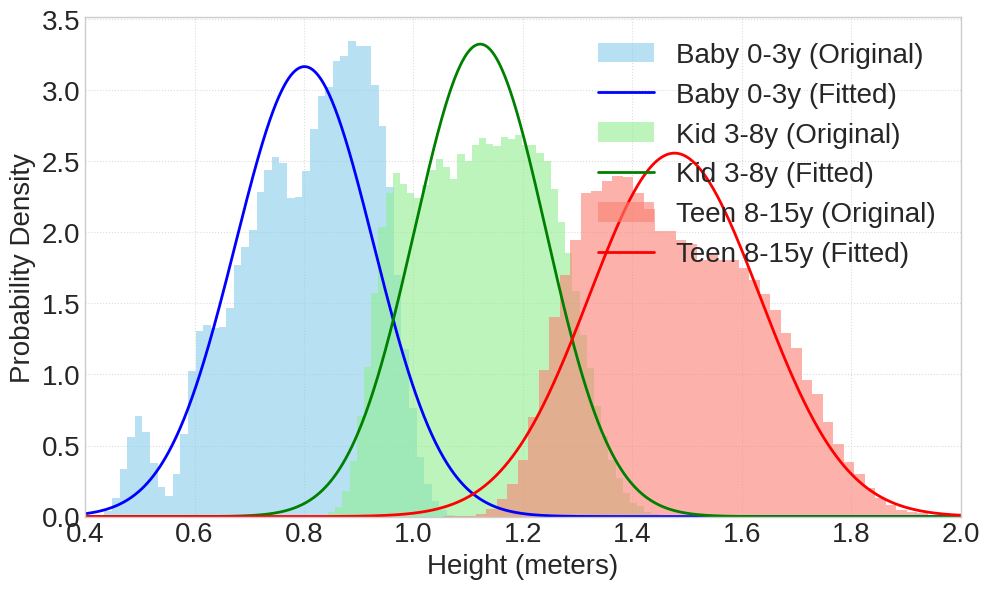}
    \caption{Height Priors for Minors.}
    \label{fig:height_prior}
\end{figure}

\paragraph{Priors for Minors}
For subjects under 15 years old, we categorize them into three age groups: 0–3 years (Baby), 3–8 years (Kid), and 8–15 years (Teen). The height priors are derived from demographic growth statistics from Centers for Disease Control and Prevention (CDC) \cite{growth}. All height values are in meters.
\begin{itemize}
\item \textbf{0–3 years (Baby):} We use the statistical height distributions for infants at 0, 3, 6, 9, 12, 18, 24, 30, and 36 months of age. By aggregating the data points sampled from these monthly distributions, we fit a single composite Gaussian, resulting in a prior of $\mathcal{N}(0.801, 0.126^2)$.
\item \textbf{3–8 years (Kid):} For this age range, we treat the height distribution for each year as a distinct Gaussian. We then form a an empirical distribution from the sum of these Gaussians and approximate it with a single, unified Gaussian, yielding a prior of $\mathcal{N}(1.122, 0.120^2)$.

\item \textbf{8–15 years (Teen):} Following the same procedure, we model the mixture of annual height distributions for the Teen with a single Gaussian, resulting in a prior of $\mathcal{N}(1.477, 0.156^2)$.

\end{itemize}
A visualization of the original height distribution and the final fitted Gaussian distribution for each group is in Fig.~\ref{fig:height_prior}.

\begin{figure}[htbp]
    \centering
    \includegraphics[width=0.9\linewidth]{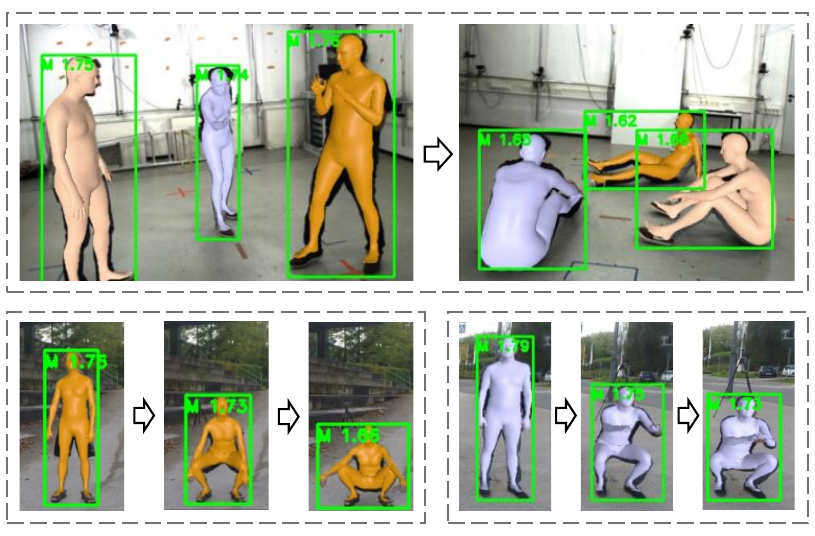}
    \caption{Illustration of height estimation bias in CameraHMR. The estimated heights are shown in green (in meters).}
    \label{fig:abnormal_height}
\end{figure}

\paragraph{Priors for Adults}
For subjects over 15 years old, we adopt a hybrid approach for their height priors. Models like CHMR (as its pGT generation method CamSMPLify) often underestimate a person's height in non-standing poses (Fig.~\ref{fig:abnormal_height}) in their effort to achieve a more accurate 2D keypoint reprojection. This can bias the scene's scale. To counteract this, we combine the model's prediction with a gender-specific statistical prior. Let $\hat{h}_{\text{CHMR}}$ be the initial height predicted by CHMR for an adult. We leverage demographic data \cite{height} which models male height as $\mathcal{N}(1.784, 0.076^2)$ and female height as $\mathcal{N}(1.647, 0.071^2)$. We denote the mean and standard deviation for the person's detected gender as $\mu_{\text{gender}}$ and $\sigma_{\text{gender}}$, respectively. The final height prior for an adult is then defined as:
$\mu_i = \frac{\hat{h}_{\text{CHMR}} + \mu_{\text{gender}}}{2} $, $\sigma_i^2 = \sigma_{\text{gender}}^2$. This formulation anchors the height estimate by averaging the model's prediction with a reliable demographic mean, while retaining the variance from the population statistics to account for natural diversity. The effectiveness of this hybrid prior is validated in our ablation study on the MuPoTS dataset (Sec~\ref{sec:appendix_ablations}).

\section{DTO-Humans}
\begin{figure}
    \includegraphics[width=\linewidth]{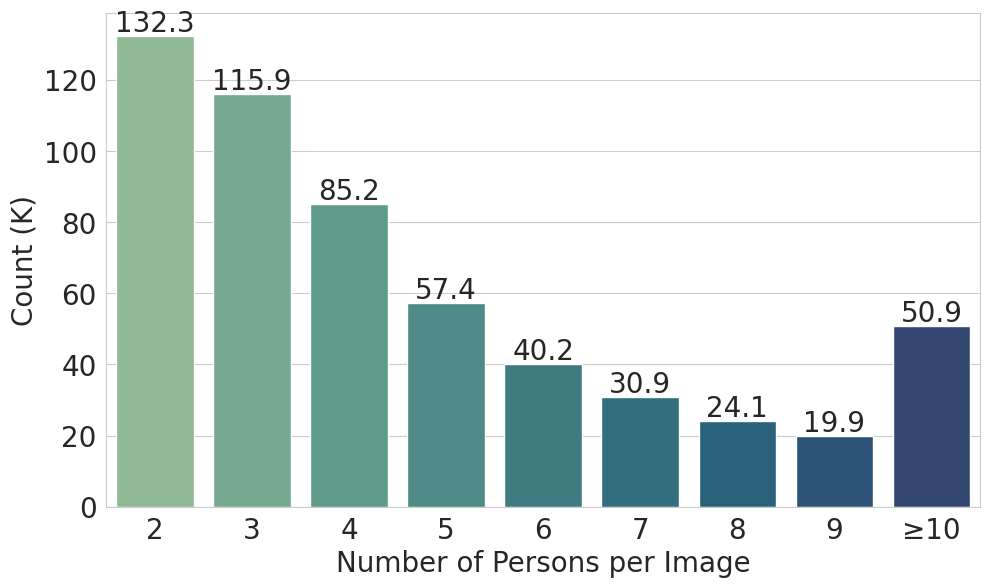}
    \caption{Number of persons per image in DTO-Humans.}
    \label{fig:dto-num-p}
\end{figure}

\begin{figure}
    \includegraphics[width=\linewidth]{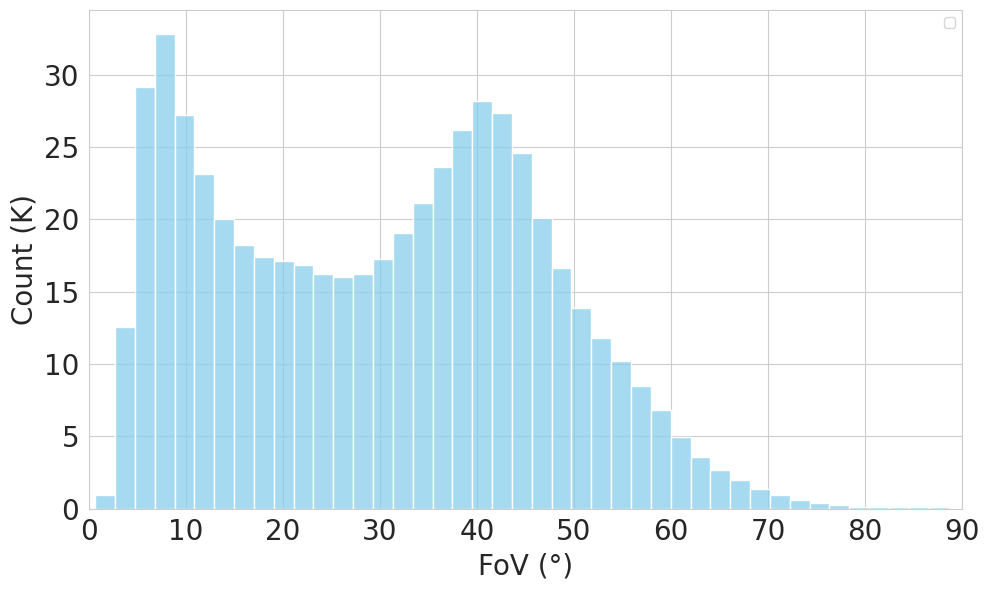}
    \caption{FoV distribution in DTO-Humans}
    \label{fig:dto-fov}
\end{figure}

The DTO-Humans dataset is curated to reflect challenging, real-world conditions. As shown in Figure \ref{fig:dto-num-p}, the dataset features a rich distribution of scenes with varying numbers of people. A large number (50.9K) of the images contain ten or more individuals, providing ample data with crowd scene consistency. Furthermore, Figure \ref{fig:dto-fov} illustrates that the dataset encompasses a wide distribution of camera FoV. This diversity in both population density and camera parameters makes DTO-Humans a valuable training dataset for robust 3D human scene understanding.

\begin{figure}
    \includegraphics[width=\linewidth]{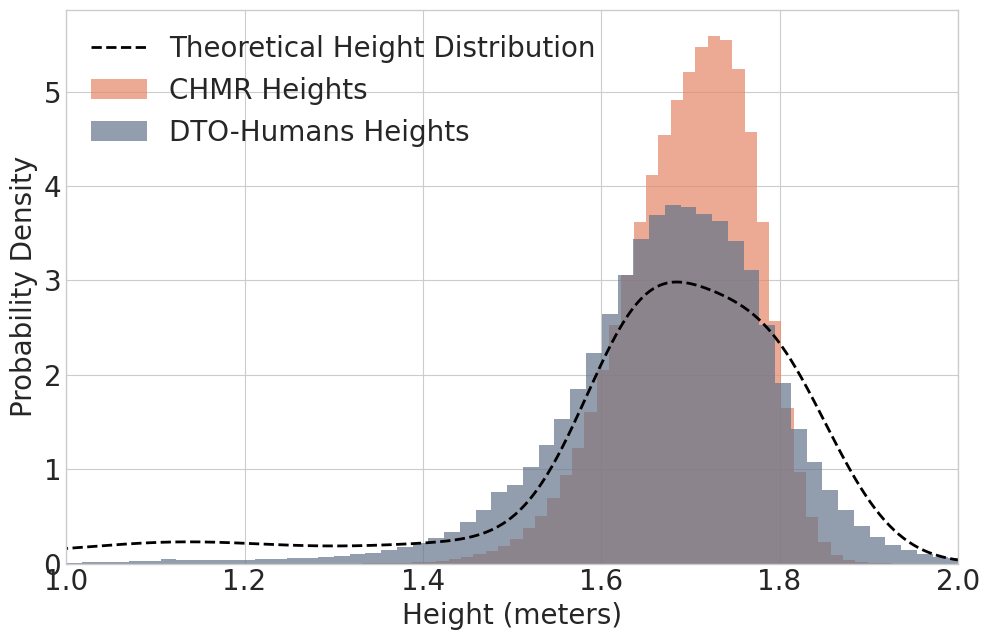}
    \caption{Comparison of Height Distributions.}
    \label{fig:dto-height}
\end{figure}

The height distribution of our DTO-Humans dataset demonstrates human size diversity derived from DTO framework, as illustrated in Fig.~\ref{fig:dto-height}. A direct comparison with the CHMR's \cite{camerahmr} version of 4D-Humans dataset reveals that the latter's height estimations are heavily concentrated in a narrow band (1.6m to 1.8m). In contrast, DTO-Humans provides a much broader and more realistic spectrum of human heights, effectively capturing individuals who are shorter or taller than this typical range.

We also analyze our dataset's distribution against a theoretical height distribution modeling a population with a uniform age distribution. While DTO-Humans aligns more closely with this ideal distribution than the baseline, the deviation caused by under-representation of non-adults remains. Specifically, minors constitute approximately 6.2\% of the people in our dataset. This is lower than the 18.8\% (i.e., 15/80) proportion that would be expected in a uniform demographic sample. Consequently, while DTO-Humans offers a broader representation of real-world height diversity, its composition still reflects the common challenge of sourcing images with a high prevalence of kids.

The visualization of samples in DTO-Humans is presented in Fig.~\ref{fig:dto-vis1}, ~\ref{fig:dto-vis2}, ~\ref{fig:dto-vis3}, ~\ref{fig:dto-vis4}.

% More vis
\twocolumn[
\begin{center}
    {\includegraphics[width=\linewidth]{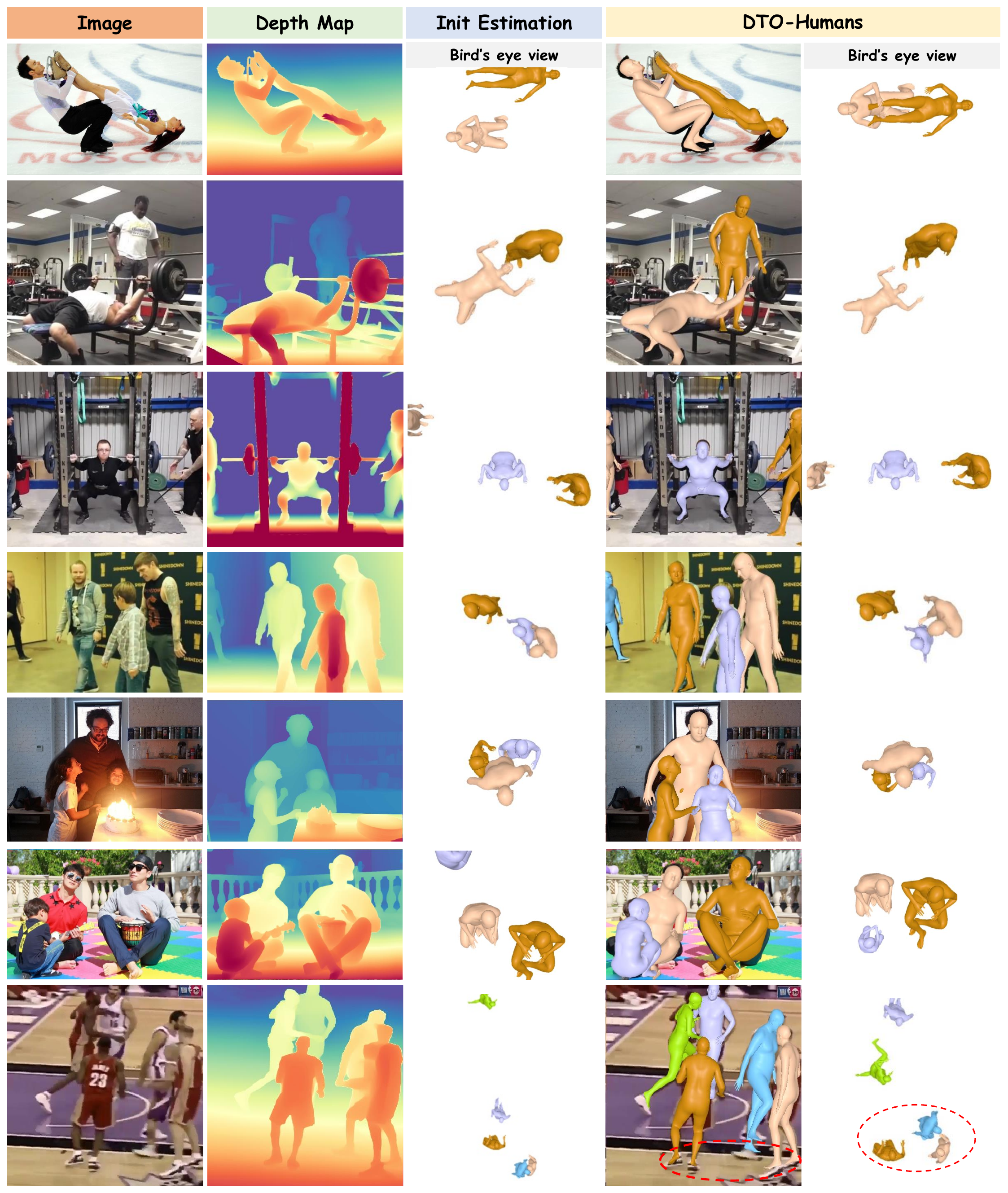}}
    \captionof{figure}{Visualization of samples in DTO-Humans}
    \label{fig:dto-vis1}
\end{center}
]

\twocolumn[
\begin{center}
    {\includegraphics[width=\linewidth]{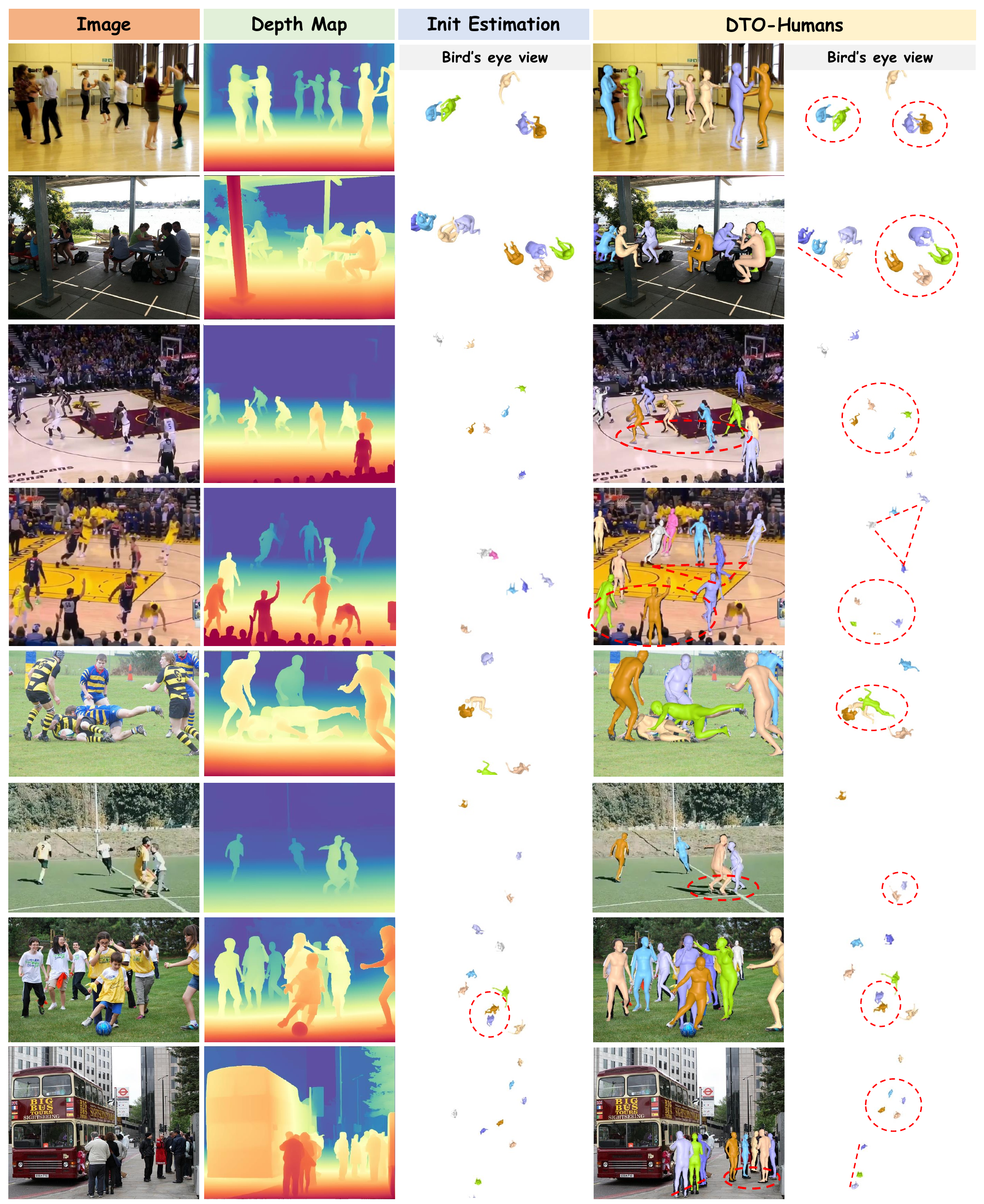}}
    \captionof{figure}{Visualization of samples in DTO-Humans}
    \label{fig:dto-vis2}
\end{center}
]

\section{MA-HMR}
\label{sec:appendix_mahmr}
\subsection{Datasets}

As mentioned in the main paper, our method is trained and evaluated on multiple publicly available datasets. Here, we provide additional details regarding their composition and usage in our experiments.

AGORA \cite{agora} is a large‑scale synthetic collection built from 4,240 high‑quality textured human scans (including 257 child scans) rendered in scenes containing 5‑15 persons per image, resulting in approximately 14K training images and 173K individual person crops. BEDLAM \cite{bedlam} is likewise synthetic, featuring monocular RGB videos of full‑body humans with ground‑truth 3D body parameters; it covers diverse body shapes, skin tones, clothing and motion in realistic scenes. In our experiments we use the training splits of both AGORA and BEDLAM (with SMPL annotations) for training MA-HMR, leveraging the broad variation of synthetic data before fine‑tuning on real‑world data.

4D‑Humans dataset \cite{4dhumans} is a large‑scale in‑the‑wild training collection that incorporates images culled from diverse sources such as InstaVariety \cite{instavariety}, COCO \cite{coco}, MPII \cite{mpii} and AI Challenger \cite{aic}, thereby enhancing generalization to real‑world scenes. In our work, we leverage the 4D‑Humans dataset to generate DTO‑Humans.

Relative Human \cite{bev} is an in‑the‑wild multi‑person dataset annotated with relative depth layers and age‑group labels. Each image contains several people, and annotations include the depth‑ordering of all persons: individuals whose depth difference is less than 0.3m share the same layer. In addition, age categories (adult, teenager, child, baby) are provided to help resolve height-depth ambiguity. It consists of approximately 7.6K images and over 24.8K person instances. In our work we use RH to evaluate the relative‑depth accuracy of our mesh reconstruction results.

3DPW \cite{3dpw}, CMU Panoptic \cite{panoptic} and MuPoTS \cite{mupots} datasets are three canonical multi‑person human mesh recovery benchmarks. Specifically, 3DPW provides challenging in‑the‑wild multi-person scenes with moving cameras. CMU Panoptic covers precisely captured multi‑person interactions in a controlled studio setting. MuPoTS focuses on generalization by providing a variety of realistic scenes for evaluation. In our work we adopt the standard evaluation protocols of prior work such as Multi‑HMR \cite{multihmr} and SAT‑HMR \cite{sathmr} on these benchmarks.

Hi4D \cite{hi4d} focuses on close‑interaction scenarios of two adults in prolonged physical contact. It comprises 20 subject pairs, covers 100 sequences and provides over 11K frames of 4D textured scans and SMPL annotations. As there is no official split, we follow the splitting protocol adopted in BUDDI \cite{buddi}. In our work, we leverage Hi4D to further assess the performance of our model in close‑interaction scenes.

\subsection{Parameter Overhead}
The proposed MA-HMR introduces a minimal parameter overhead compared to the baseline SAT-HMR. The addition of a 4-layer MLP for camera FoV regression slightly increases the total parameter count from 221.9M to 223.7M (+0.8\%), demonstrating the efficiency of our approach. 

\subsection{Hyperparameters}
The weighting coefficients for each term in MA-HMR's training loss function, $\lambda_{\text{map}}$,
$\lambda_{\text{depth}}$, $\lambda_{\text{pose}}$, $\lambda_{\text{shape}}$, $\lambda_{\text{j3ds}}$, $\lambda_{\text{j2ds}}$, $\lambda_{\text{box}}$, $\lambda_{\text{det}}$, $\lambda_{\text{fov}}$, and $\lambda_{\text{rm}}$, are set to 4.0,
0.5, 5.0, 3.0, 8.0, 40.0, 2.0, 1.0, 0.5 and 0.5, respectively.

\subsection{Qualitative Results}
In Fig.~\ref{fig:cmp_multi}, we compare with Multi-HMR under different FoV settings. We then compare with SAT-HMR on 3DPW and CMU Panoptic (Fig.\ref{fig:3dpw}, \ref{fig:panoptic}), revealing our pose accuracy and spatial consistency. Fig.~\ref{fig:rh_vis} presents results on Relative Human, where large variations in subject distance and scale further demonstrate the robustness of our approach.

\section{More Ablations}
\label{sec:appendix_ablations}

\subsection{Ablation on Gender-Aware Height Prior}

% As illustrated in Figure~\ref{fig:abnormal_height}, monocular mesh recovery methods like CameraHMR (CHMR) tend to underestimate a person's canonical height in non-standing poses (e.g., sitting or crouching) in their effort to achieve a more accurate 2D keypoint reprojection. Our prior is designed to counteract this bias by anchoring the height estimate to a more realistic demographic mean. We conduct an ablation study to validate the effectiveness of our gender-aware height prior.

\begin{table}[h]
    \centering
    \caption{Ablation study on MuPoTS evaluating our DTO components. S: segmentation-enhanced inference. D': optimization with gender-agnostic height prior. D: optimization with with gender-aware height prior. X: intra-human depth scale constraints. }
    \begin{tabular}{l c}
        \toprule
        Method & MPJPE $\downarrow$ \\
        \midrule
        CHMR & 89.5 \\
        CHMR+S & 87.2 \\
        CHMR+S+D' & 87.1 \\
        CHMR+S+D'+X & 86.7 \\
        \textbf{CHMR+S+D+X} & \textbf{86.3} \\
        \bottomrule
    \end{tabular}
    \label{tab:ablation_gender}
\end{table}

\twocolumn[
\begin{center}
    {\includegraphics[width=\linewidth]{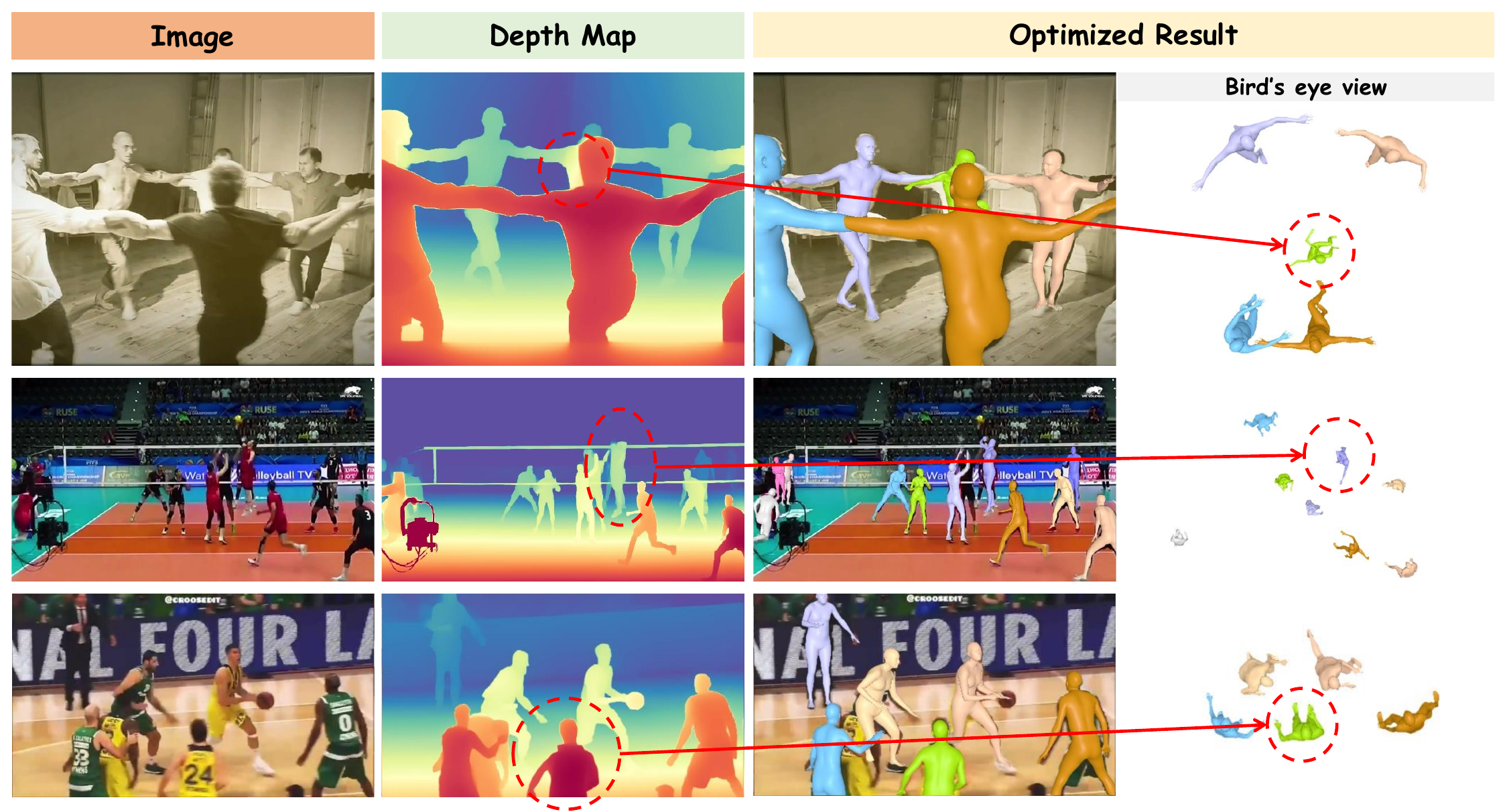}}
    \captionof{figure}{Examples of DTO failure cases, primarily caused by inaccurate relative depth maps.}
    \label{fig:limitation}
\end{center}
]

We evaluate gender-aware height prior on MuPoTS, and the quantitative results are shown in Table~\ref{tab:ablation_gender}. The baseline CHMR is progressively improved by adding segmentation-enhanced inference (+S). We then compare our full model (+D+X), which uses the gender-aware prior, against a variant where DTO is applied without this prior (+D'+X).

The introduction of the gender-aware prior (+D) further reduces the MPJPE from 86.7 to 86.3. This confirms that guiding the optimization with a statistically-grounded height mean effectively mitigates the model's underestimation bias, leading to a more accurate final 3D reconstruction.

\begin{table}[h]
    \centering
    \caption{Performance evaluation of different Depth Anything v2 (DAv2) backbones on Relative Human.}
    \begin{tabular}{l c c}
        \toprule
        HMR Model & Depth Model & PCDR$_{0.2}$ (all) $\uparrow$ \\
        \midrule
        InstaHMR \cite{instahmr} & / & 72.83 \\
        CHMR & DAv2-S & 72.93 \\
        CHMR & DAv2-B & 73.74 \\
        CHMR & DAv2-L & \textbf{74.16} \\
        \bottomrule
    \end{tabular}
    \label{tab:ablation_depth_model}
\end{table}

\subsection{Ablation on Depth Model}
To select the optimal depth estimation model for our framework, we evaluated different backbones for the Depth Anything v2 model on the Relative Human dataset. As shown in Table~\ref{tab:ablation_depth_model}, we tested ViT-Small, Base and Large.

It is noteworthy that even using the smallest DAv2-S backbone yields a PCDR score of 72.93, which surpasses the state-of-the-art performance of InstaHMR (72.86), indicating the low entry barrier and high efficiency of our DTO framework. As DAv2-L offers the best performance for our task, we selected it as our default depth model.

\section{Limitations}
\label{sec:limitations}
The limitation of our DTO framework lies in its dependency on the accuracy of the upstream relative depth estimation model. Errors or ambiguities in the generated depth map can propagate, leading to implausible scene reconstructions. Fig.~\ref{fig:limitation} illustrates several typical failure modes. Firstly, severe occlusion can cause the depth model to assign a foreground depth value to a background person, making DTO incorrectly scale them down (top row). In scenes with ambiguous depth cues, such as an athlete jumping mid-air, the depth model may resort to flawed heuristics (e.g., higher in image means farther away), misplacing the person in the scene (middle row). Lastly, partial visibility without a clear ground plane can confuse the depth model, leading it to incorrectly flatten the relative depths of multiple people, which DTO then inherits (bottom row). On the other hand, these cases underscore that future advancements in monocular depth estimation can directly enhance the accuracy and robustness of our DTO framework, thus lead to better 3D human scene understanding.

\twocolumn[
\begin{center}
    {\includegraphics[width=\linewidth]{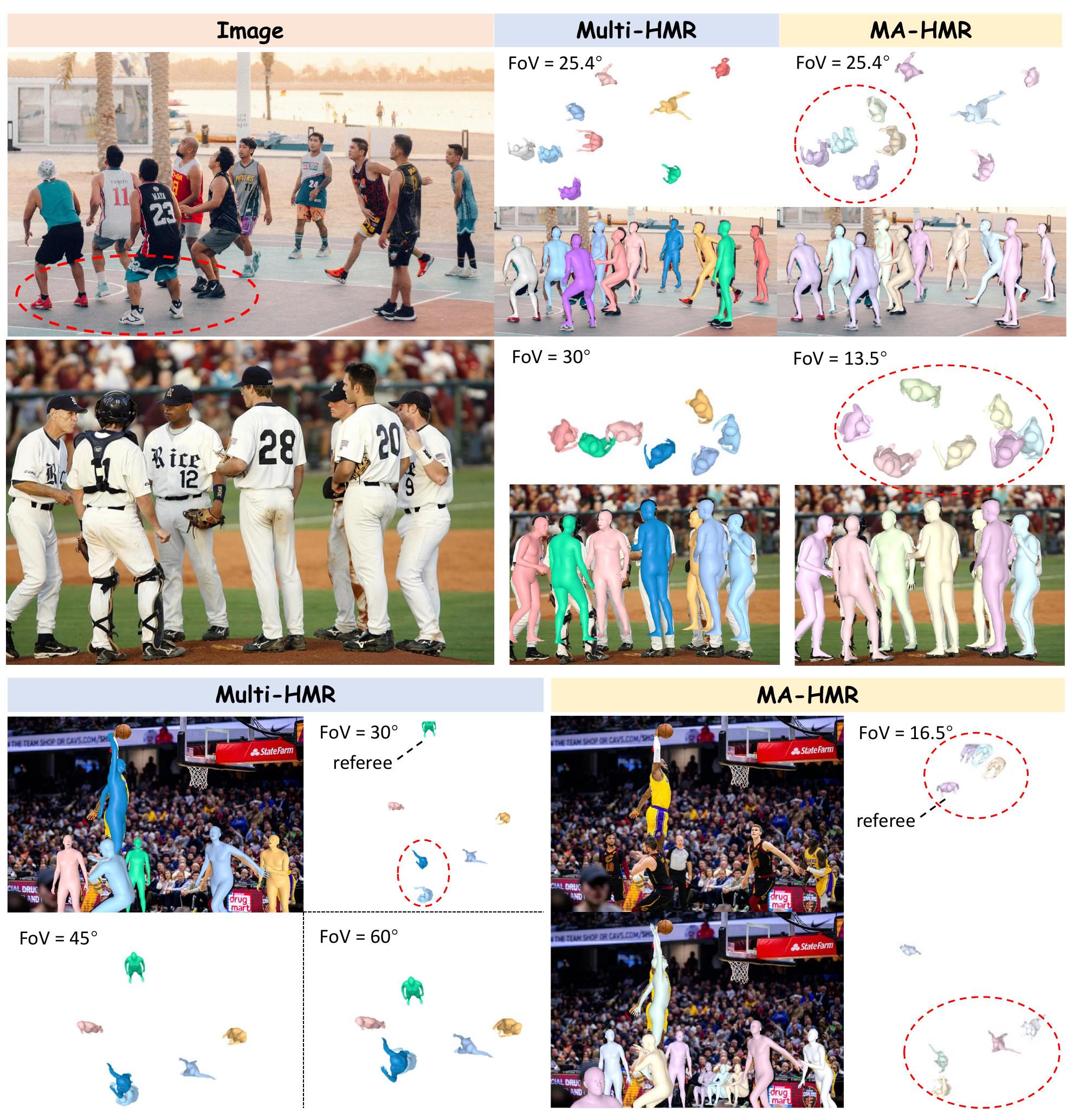}}
    \captionof{figure}{Qualitative Comparison with Multi-HMR \cite{multihmr}. Multi-HMR supports an optional FoV input to recover human meshes in camera space. For comparison, we test Multi-HMR in different FoV settings. (The definition of the displayed FoV here follows Multi-HMR, which is the larger of the vertical FoV and horizontal FoV). Notably, in the bottom cases, our MA-HMR successfully handles both the near-field action (the dunk over a defender) and the far-field subjects (the referee outside the three-point line and the spectators). In contrast, when we narrow Multi-HMR's input FoV for correctly placing the distant referee, its reconstruction of the near-field dunk deteriorates.}
    \label{fig:cmp_multi}
\end{center}
]

\twocolumn[
\begin{center}
    {\includegraphics[width=1\linewidth]{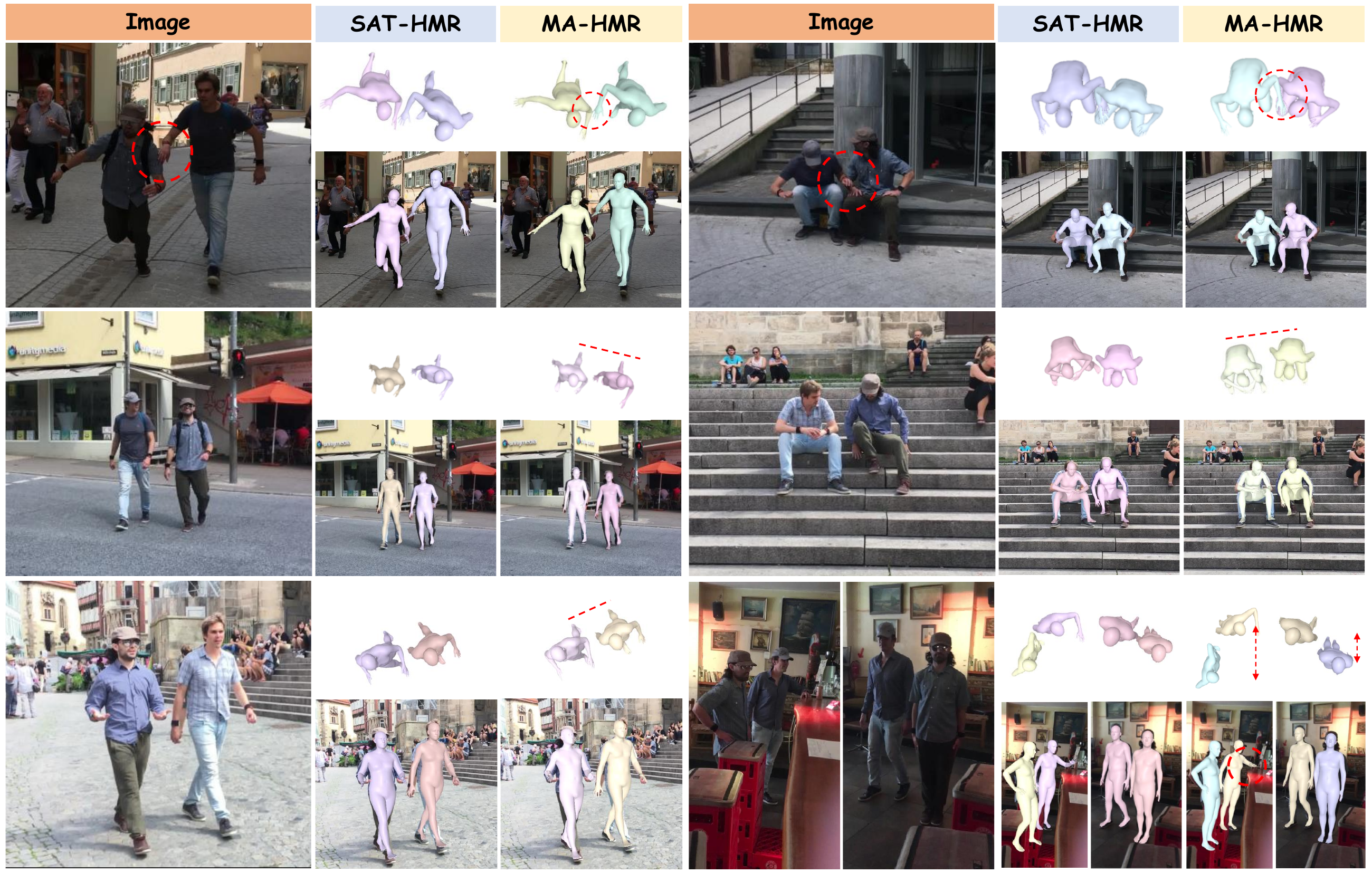}}
    \captionof{figure}{Qualitative Results of MA-HMR on 3DPW, with comparison to SAT-HMR \cite{sathmr}}
    \label{fig:3dpw}
\end{center}
\begin{center}
    {\includegraphics[width=1\linewidth]{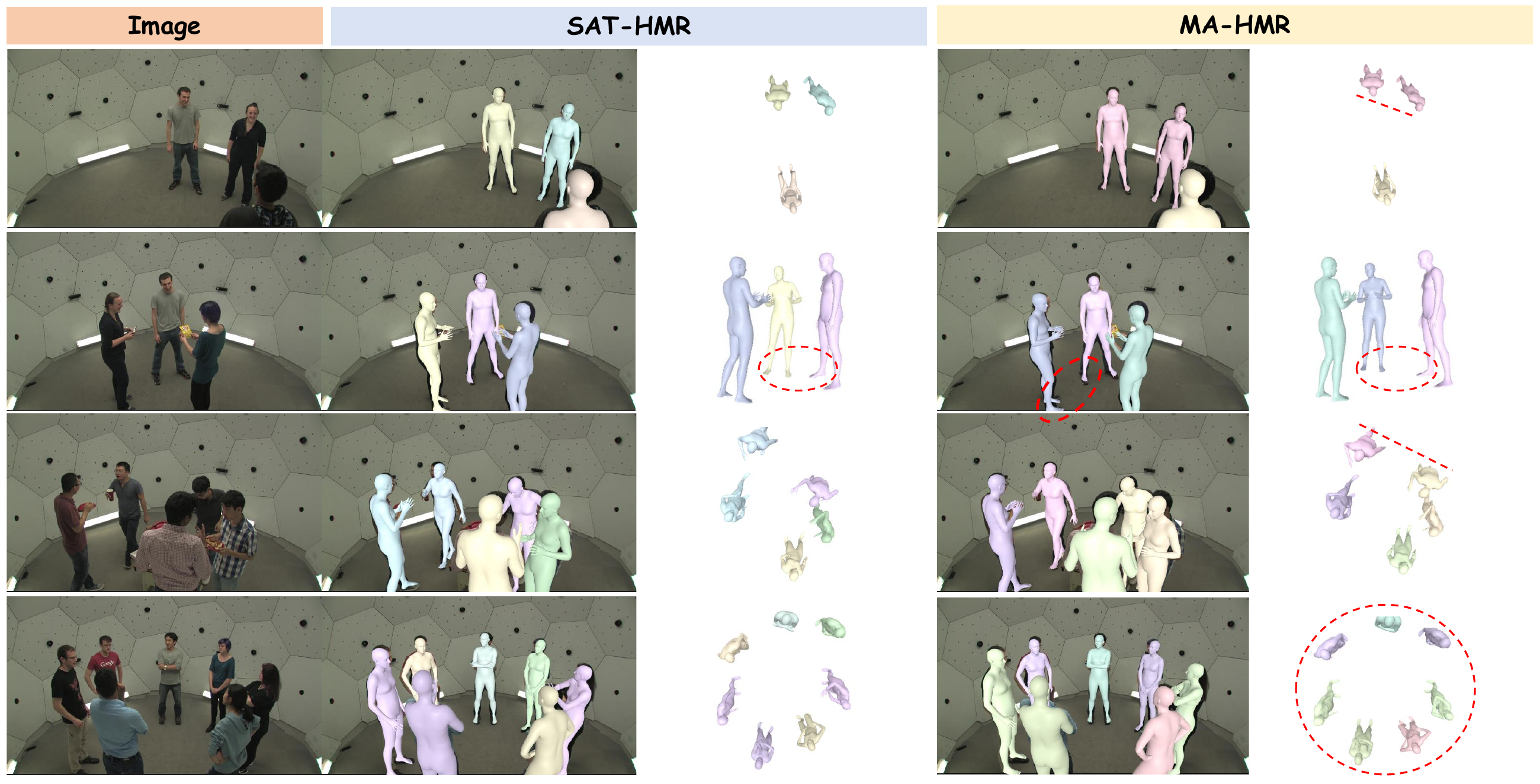}}
    \captionof{figure}{Qualitative Results of MA-HMR on CMU Panoptic, with comparison to SAT-HMR \cite{sathmr}}
    \label{fig:panoptic}
\end{center}
]

\twocolumn[
\begin{center}
    {\includegraphics[width=\linewidth]{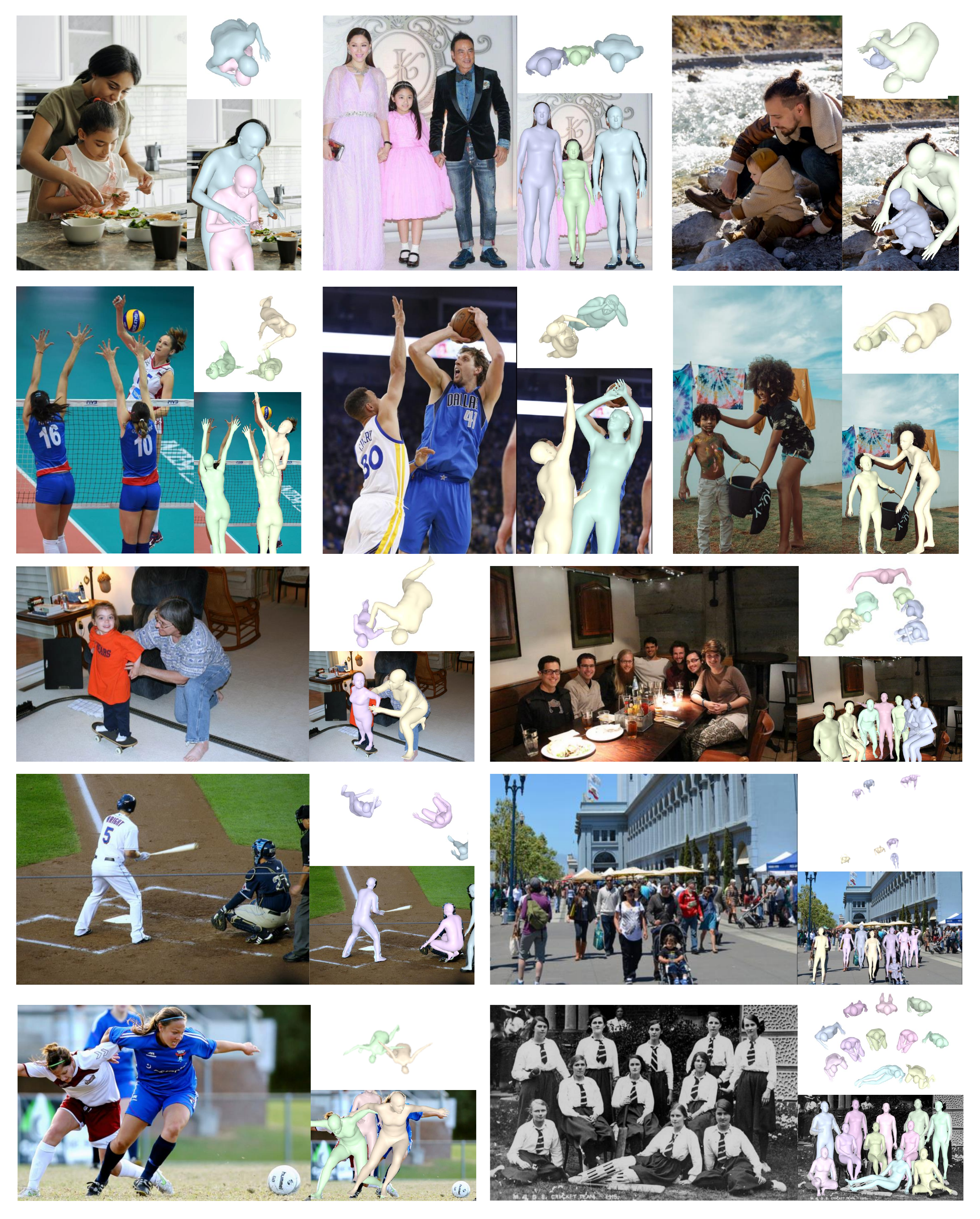}}
    \captionof{figure}{Qualitative Results of MA-HMR on Relative Human.}
    \label{fig:rh_vis}
\end{center}
]

\twocolumn[
\begin{center}
    {\includegraphics[width=\linewidth]{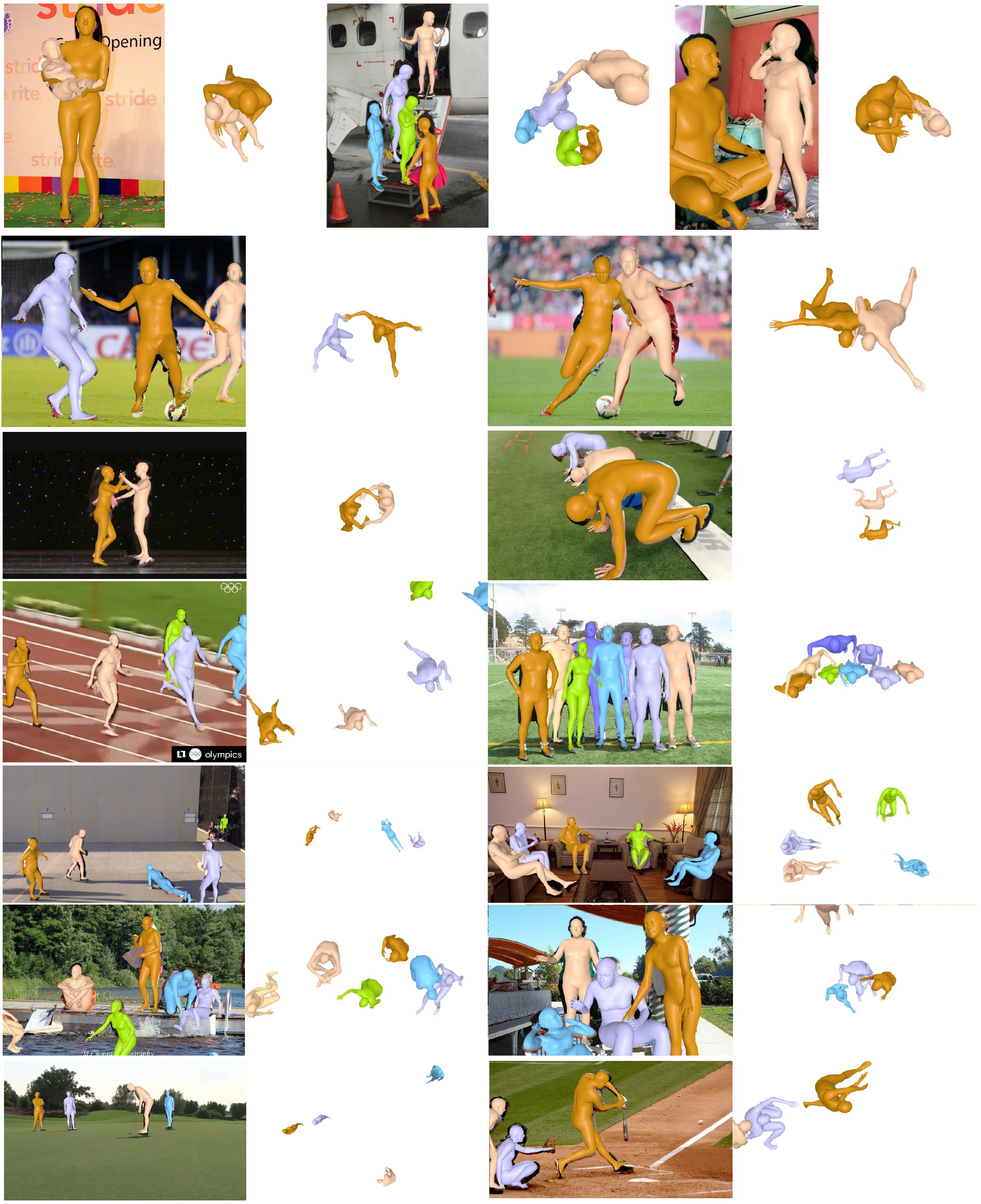}}
    \captionof{figure}{More visualization of samples in DTO-Humans.}
    \label{fig:dto-vis3}
\end{center}
]

\twocolumn[
\begin{center}
    {\includegraphics[width=0.95\linewidth]{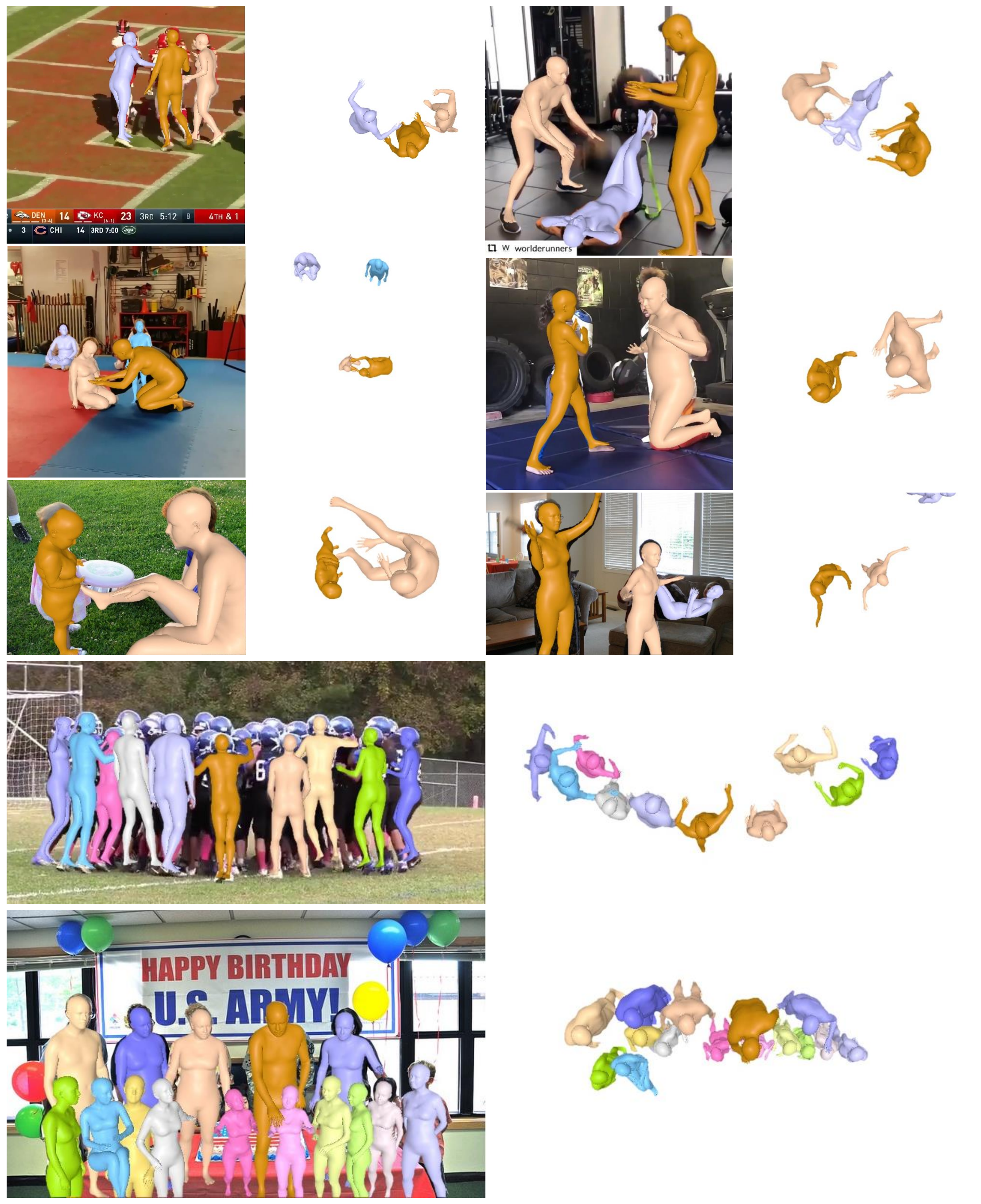}}
    \captionof{figure}{More visualization of samples in DTO-Humans.}
    \label{fig:dto-vis4}
\end{center}
]

\end{document}